\documentclass{article}
\PassOptionsToPackage{numbers}{natbib}
\usepackage[main, final]{neurips_2026}
\usepackage[utf8]{inputenc} 
\usepackage[T1]{fontenc}    
\usepackage{hyperref}       
\usepackage{url}            
\usepackage{booktabs}       
\usepackage{amsfonts}       
\usepackage{nicefrac}       
\usepackage{microtype}      
\usepackage[table]{xcolor}  

\usepackage[most]{tcolorbox}
\usepackage{fvextra}
\usepackage{graphicx}
\usepackage{subcaption}
\usepackage{multirow}
\usepackage{enumitem}       
\usepackage{bm}
\usepackage{amsmath}
\usepackage{amssymb}
\usepackage{mathtools}
\usepackage{algorithm}
\usepackage{algorithmic}
\usepackage{wrapfig}
\usepackage{amsthm}
\usepackage{placeins}
\usepackage{needspace}
\usepackage{pifont}
\theoremstyle{plain}
\newtheorem{theorem}{Theorem}[section]

\theoremstyle{definition}
\newtheorem{definition}[theorem]{Definition}

\theoremstyle{remark}
\newtheorem{remark}[theorem]{Remark}
\newcommand{\partitle}[1]{\vspace{0.3em} \noindent \textbf{#1.}}

\title{Ask in the Crowd: Differentially Private LLM Inference via Dummy-Augmented Shuffling}

\author{
Zhihao Liu$^{1}$\thanks{These authors contributed equally to this work.} \quad
Zixiong Guo$^{1}$\footnotemark[1] \quad
Shuo Shao$^{1}$ \quad
Yu He$^{1}$ \quad
Wenli Wang$^{1}$ \And
Meihui Chen$^{2}$\thanks{Corresponding authors.} \quad
Di Wang$^{3}$ \quad
Yuke Hu$^{3}$\footnotemark[2] \\[1mm]
$^{1}$Zhejiang University \quad
$^{2}$Xiaomi Corporation \quad
$^{3}$King Abdullah University of Science and Technology \\[1mm]
\texttt{\{zhihao\_liu, guozixiong, shaoshuo\_ss, yuherin, wangwenli\}@zju.edu.cn} \\
\texttt{chenmeihui@xiaomi.com} \quad
\texttt{\{di.wang, yuke.hu\}@kaust.edu.sa}
}

\begin{document}

\maketitle

\begin{abstract}
Large language models are increasingly deployed as LLM-as-a-Service (LLMaaS), where user queries are sent to remote LLMs for inference, raising severe privacy concerns about the exposure of sensitive user inputs. Differentially private inference mitigates these risks by injecting noise into input embeddings. However, existing methods struggle to balance privacy and utility, as stronger privacy protection often leads to substantial accuracy degradation in LLM inference.

In this work, we propose DP-SPD, a differentially private LLM inference framework that protects user queries by hiding the real query among client-generated dummy queries. All queries are randomly shuffled before being sent to the server, preventing direct identification of the user's true intent. 
To improve inference efficiency, we exploit the shared-prefix structure among queries and reuse transformer key--value (KV) caches during decoding.
We analyze the privacy guarantee of DP-SPD and evaluate its utility on classification and generation tasks, as well as its privacy protection under QIA, EIA, and GPT-based attacks. 
Experimental results show that our method achieves strong inference-time privacy protection and near-plaintext utility while improving inference efficiency through shared-prefix KV-cache reuse.

\end{abstract}

\section{Introduction}

Large language models (LLMs) are increasingly offered as LLM-as-a-Service (LLMaaS) and deployed on cloud-based platforms \cite{brown2020language,ouyang2022training,chowdhery2022palm,qwen2offcial2025}, where user queries are transmitted to remote servers for inference, enabling scalable and high-quality language understanding. However, this deployment paradigm also raises serious privacy concerns \cite{chen2025survey,bai2025safeguarding}: user inputs may contain private information, such as medical records, health conditions, financial details, location traces, or private communications, which are directly exposed to the LLM service provider. Such exposure risks data leakage, misuse, or memorization by the model, making privacy-preserving inference essential for safe and trustworthy LLMaaS deployment.

However, existing approaches to privacy-preserving LLM inference face a difficult trade-off among privacy, utility, and efficiency. Cryptography-based methods such as secure computation \cite{rathee2024mpc,zeng2025mpcache} and homomorphic encryption \cite{de2024privacy,ruoyan2025practical} which provide strong privacy guarantees by design, as the LLM server never observes raw user inputs. However, their substantial computational and communication overhead often limits their practicality in large-scale LLMaaS deployments. 
Perturbation-based DP methods \cite{du2023dp,mai2024split,tong2025inferdpt,lin2025obfuslm,thareja2025dp,liu2026dpmind} incur lower computational overhead, but injecting random perturbations into tokens or embeddings inevitably degrades output quality \cite{tong2025inferdpt,thareja2025dp}, while denoising-based methods introduce additional inference-time cost \cite{mai2024split}. As a result, existing DP LLM inference approaches still face a practical challenge in balancing privacy, utility, and efficiency.

This limitation motivates us to move beyond relying solely on stronger perturbation and instead introduce anonymity into private LLM inference. Specifically, we explore client-side dummy augmentation, where the real query is hidden among multiple plausible dummy queries before being sent to the server.
In this way, the server no longer observes a single identifiable request, but a shuffled set of hard-to-distinguish candidate queries.
However, unlike classical shuffle DP, LLM inference usually involves a single user query.
Therefore, the client must locally construct the anonymity set rather than relying on reports from other independent users.
Such client-side construction introduces two challenges: \ding{172} dummy queries must be contextually plausible enough to avoid being filtered by the server, and \ding{173} dummy-augmented inference must remain efficient enough for practical use.

In this paper, we aim to improve the ``privacy--utility--efficiency'' trade-off for differentially private LLM inference. We propose a shuffle-based privacy-preserving inference framework DP-SPD that hides the real user query among dummy queries. DP-SPD first selects and privatizes a shared prefix in the continuous embedding space under metric DP. A lightweight client-side continuation model then generates prefix-consistent dummy suffixes from the privatized prefix, providing stronger camouflage for the real continuation. Finally, the real and dummy inputs are randomly shuffled on the client side before being sent to the server. Furthermore, by exploiting the shared-prefix structure together with transformer key--value (KV) cache reuse, DP-SPD substantially reduces the overhead of dummy-augmented inference. Our main contributions are summarized as follows:

\begin{itemize}[leftmargin=*, labelindent=0.2em]
    \item We propose a shuffle-based privacy-preserving inference framework DP-SPD that combines prefix-consistent dummy suffix generation with client-side shuffling. By sharing prefixes across real and dummy queries, DP-SPD enables KV-cache reuse and reduces redundant inference overhead.
    
    \item We provide rigorous privacy and correctness analyses for the framework, formally establishing its guarantees under $d_\chi$-privacy and proving that the client can reliably identify the output corresponding to the real input.
    
    \item We conduct extensive experiments on multiple models and tasks to evaluate utility and inference efficiency. We also evaluate privacy leakage under query identification attack, embedding inversion attack and GPT-based attack to demonstrate that DP-SPD achieves a stronger privacy--utility--efficiency trade-off than existing perturbation-based baselines.
\end{itemize}

\section{Background and Related Work}

\subsection{Inference-Time Privacy Risks in LLMaaS}

Large Language Models (LLMs) are increasingly deployed as LLM-as-a-Service (LLMaaS) \cite{brown2020language,ouyang2022training,chowdhery2022palm,qwen2offcial2025, hu2024eraser}, where users access cloud-hosted models through black-box APIs without visibility into model parameters or inference internals. This deployment paradigm enables scalable and cost-efficient access to foundation models, but fundamentally shifts inference to a centralized setting. However, LLMaaS requires transmitting user prompts verbatim to centralized servers, exposing sensitive inference-time inputs to the service provider. Prior work \cite{zhang2024privacyasst,dong2025depth} has demonstrated that such inputs may be persistently logged, implicitly memorized by models, or indirectly leaked through system-level channels and auxiliary artifacts, raising serious concerns about the privacy of user interactions. As a result, such privacy risks are particularly concerning for real-world LLMaaS deployments involving sensitive or proprietary content. 

\subsection{Differentially Private LLM Inference}

To address inference-time privacy risks in LLMaaS, perturbation-based mechanisms have emerged as a practical approach for privacy-preserving LLM inference. DP-Forward \cite{du2023dp} perturbs input embeddings and forwards the noisy representations directly to the LLM for inference. Split-and-Denoise (SnD) \cite{mai2024split} further uses an auxiliary denoising module to recover semantic information from the perturbed embeddings. In contrast, InferDPT \cite{tong2025inferdpt} introduces a prompt-level local differential privacy mechanism combined with response extraction strategies. ObfusLM \cite{lin2025obfuslm} operates in the embedding space and employs obfuscation and shuffling to defend against inversion and query identification attacks. Despite these advances, the inherent characteristics of LLM inference, most notably the \textbf{single-query} execution paradigm, fundamentally constrain the effective use of differential privacy amplification. Beyond this core challenge, existing methods suffer from additional drawbacks, including reliance on extra denoising models and increased semantic distortion caused by discrete token perturbation. As a result, achieving a balanced ``privacy–utility–efficiency'' trade-off for differentially private LLM inference remains an open problem.

\section{Preliminary}

\subsection{Differential Privacy}
Differential privacy (DP) \cite{dwork2006our} is a standard formal notion of data privacy, defined as follows:

\begin{definition}[($\epsilon, \delta$)-Differential Privacy]
A randomized mechanism $M$ with domain $D$ and range $R$ preserves $(\epsilon, \delta)$-differential privacy if for any two neighboring datasets $D, D' \in D$ and for any subset $S \subseteq R$, the following inequality holds:
\begin{equation}
\Pr[M(D) \in S] \le e^\epsilon \Pr[M(D') \in S] + \delta
\end{equation}
where $\epsilon$ is the privacy budget and $\delta$ is the failure probability.
\end{definition}

Local differential privacy (LDP) is a client-side privacy model in which each user locally perturbs their input before sending it to an untrusted server.

\begin{definition}[$\epsilon$-Local Differential Privacy]
A randomized mechanism $M : \mathcal{X} \rightarrow \mathcal{Y}$ satisfies $\epsilon$-local differential privacy if for any two inputs $x, x' \in \mathcal{X}$ and for any measurable set $S \subseteq \mathcal{Y}$,
\begin{equation}
\Pr[M(x) \in S] \le e^{\epsilon} \Pr[M(x') \in S].
\end{equation}
\end{definition}

\begin{definition}[$d_\chi$-Privacy]
Let $(\mathcal{X}, d_\chi)$ be a metric space where $d_\chi$ denotes a distance metric (e.g., Euclidean distance). A randomized mechanism $M: \mathcal{X} \to \mathcal{Y}$ satisfies $\epsilon$-$d_\chi$-privacy if for any $x, x' \in \mathcal{X}$ and any $S \subseteq \mathcal{Y}$:
\begin{equation}
    \Pr[M(x) \in S] \le e^{\epsilon \cdot d_\chi(x, x')} \Pr[M(x') \in S].
\end{equation}
\end{definition}
Given a tokenized query sequence $X=[x_1,\ldots,x_n]$, the embedding layer maps it to $\mathbf{X}=[\bm{x}_1,\ldots,\bm{x}_n] \in \mathbb{R}^{n\times d}$ via an embedding matrix $\mathbf{E}\in\mathbb{R}^{|\mathcal{V}|\times d}$, where $|\mathcal{V}|$ is the vocabulary size and $d$ is the embedding dimension. Under the $d_\chi$-privacy metric with parameter $\epsilon$ and $L_2$ distance, privacy is enforced by adding Laplacian noise $\bm{z}$ with density proportional to $\exp(-\epsilon\|z\|)$ \cite{wu2017bolt}. Specifically, the noise is sampled as $\bm{z} = lv$, where $l \sim \Gamma(d, 1/\epsilon)$ and $v$ is uniformly drawn from the unit ball $B^d$, yielding the privatized embedding $M(\bm{x}_t) = \bm{x}_t + \bm{z}$.

Furthermore, to mitigate the utility degradation caused by local noise, we operate under the shuffle model of differential privacy \cite{bittau2017prochlo, erlingsson2019amplification, li2023privacy_dump}.

\begin{definition}[Shuffle Model]
Consider $k$ users, each holding a private input $x_i$. The Shuffle Model consists of a local randomizer $\mathcal{R}$ and a trusted shuffler $\mathcal{A}$. Each user applies $y_i = \mathcal{R}(x_i)$, and the shuffler applies a uniform random permutation $\pi$ to the batch of messages. The server observes the multiset:
\begin{equation}
    \mathcal{M}_{shuffle}(x_1, \dots, x_k) = \{ y_{\pi(1)}, y_{\pi(2)}, \dots, y_{\pi(k)} \}.
\end{equation}
\end{definition}

The core advantage of this model is \textit{Privacy Amplification}: shuffling provides a stronger central DP guarantee than the local guarantee of $\mathcal{R}$ alone, since the adversary cannot easily link a specific output message $y_j$ back to a specific user $i$. 

\subsection{KV-cache in LLM Inference}

To mitigate the computational redundancy of autoregressive generation, LLMs employ the \textit{KV-cache} mechanism~\cite{pope2023efficiently, luo2026shadow}. 
Rather than recomputing the Key ($K$) and Value ($V$) representations for the entire context at each decoding step, the model caches these intermediate states from previous tokens. 
At step $i$, the model computes only the current query, key, and value vectors $(q_i, k_i, v_i)$, attends $q_i$ over the cached keys and values, and then appends the new $(k_i, v_i)$ pair to the cache. 
This design substantially improves inference efficiency, at the cost of maintaining a persistent cache whose size grows linearly with the sequence length.

\section{Methodology}

We present \textbf{D}ifferentially \textbf{P}rivate \textbf{S}huffling with \textbf{P}refix-conditioned \textbf{D}ummies (DP-SPD), a privacy-preserving inference framework for LLMs. Given a user query, the client generates multiple dummy suffixes conditioned on the same prefix, locally perturbs all candidate suffixes, and randomly shuffles them before sending them to the server. The shared-prefix design further enables KV-cache reuse, reducing the inference overhead introduced by dummy augmentation.

\begin{figure}[t]
    \centering
    \includegraphics[width=0.85\linewidth]{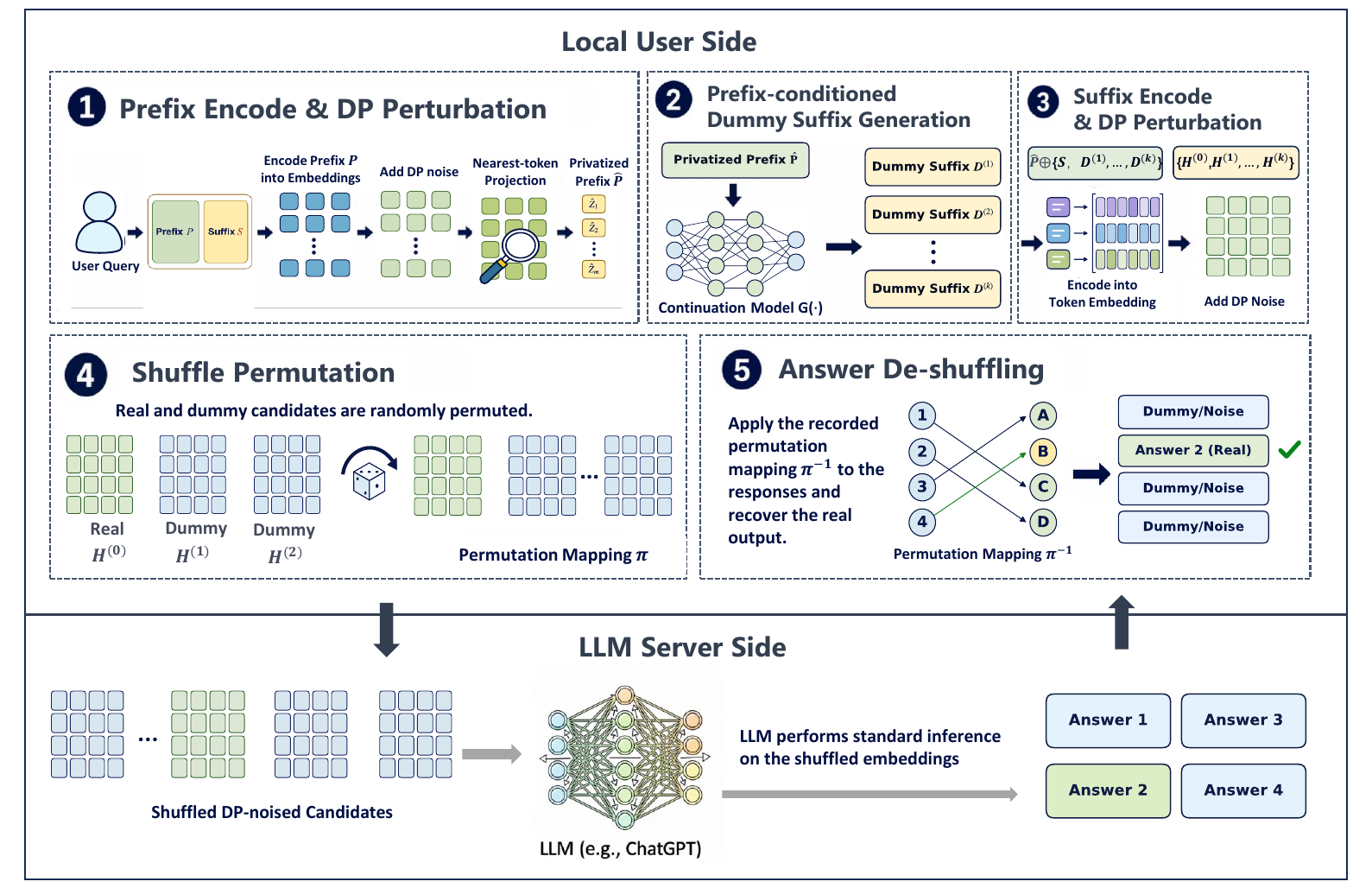}
    \caption{Full workflow of DP-SPD. The client generates prefix-consistent dummy suffixes, perturbs and shuffles them locally, and sends the shuffled noisy candidates to the server for inference.}
    \vspace{-1em}
    \label{fig:appendix_workflow}
\end{figure}

\subsection{DP-SPD Mechanism}

Our design is motivated by two properties of LLM inference that make standard privacy mechanisms difficult to apply directly.

\ding{172} Token-level differences do not faithfully reflect \textbf{semantic distance}. Queries with many token changes may remain semantically close, while small token changes can substantially alter the meaning. Thus, standard local differential privacy can require overly strong perturbation and hurt utility. We instead adopt \textbf{Metric Differential Privacy} ($d_\chi$-privacy)~\cite{chatzikokolakis2013broadening,feyisetan2020privacy}, which measures privacy loss by embedding distance and better captures semantic privacy in LLM inference.

\ding{173} Shuffling cannot be directly applied to \textbf{single-user LLM inference}. Classical privacy amplification by shuffling relies on reports collected from multiple independent users, where anonymity comes from hiding which randomized report belongs to which user. In LLM inference, however, the interaction is typically initiated by a single user. 
A natural remedy is to introduce synthetic dummy inputs on the client side, but naive dummy construction leads to a tension between privacy and efficiency. Sentence-level random sampling can create full dummy queries that are less trivially tied to the real input, but each dummy query requires a separate forward pass, substantially increasing the \textbf{inference overhead}. Token-level random sampling allows prefix sharing and \textbf{KV-cache} reuse, but such token-level dummies are often contextually or \textbf{semantically inconsistent} with the real query. As a result, the server may distinguish them using contextual consistency, semantic relevance, or prior knowledge, causing the effective anonymity set to collapse.

DP-SPD addresses these issues by constructing dummies under the same shared prefix as the real query. Prefix conditioning makes the dummy suffixes more context-compatible with the real suffix, so that the shuffled candidates form a more meaningful anonymity set. At the same time, local perturbation provides a metric-DP guarantee over each candidate, while shuffling the real and dummy suffixes further obscures which perturbed suffix corresponds to the user's true private input. Thus, DP-SPD combines metric-based local perturbation, prefix-consistent dummy construction, and client-side shuffling into a unified framework for practical privacy-preserving LLM inference.

\begin{figure}[t]
\begin{algorithm}[H]
\caption{DP-SPD Main Pipeline}
\label{alg:dummy_mechanism_new}
\begin{algorithmic}[1]
\REQUIRE Tokenized query $X=[x_1,\dots,x_n]$, continuation model $G$, embedding function $E$, projection operator $\Pi_{\mathcal V}$, dummy ratio $r$, dummy number $k$, privacy budgets $\epsilon_{\mathrm{pre}},\epsilon_{\mathrm{suf}}$, deployment mode $m$
\ENSURE Server responses $\mathcal{Y}$ and real-query output $y_{\mathrm{real}}$

\STATE $T \leftarrow \lfloor (1-r)n \rfloor$
\STATE $\mathcal{P}\leftarrow [x_1,\dots,x_T]$, \quad $\mathcal{S}\leftarrow [x_{T+1},\dots,x_n]$
\STATE $(\widetilde{\mathcal{P}}_{\mathrm{emb}},\widehat{\mathcal{P}}) \leftarrow \textsc{PrivatizePrefix}(\mathcal{P},E,\Pi_{\mathcal V},\epsilon_{\mathrm{pre}})$
\STATE $\mathcal{D}^{(1)},\dots,\mathcal{D}^{(k)} \sim G(\widehat{\mathcal{P}}, n-T)$
\STATE $H^{(0)},\dots,H^{(k)} \leftarrow \textsc{BuildCandidates}(\widetilde{\mathcal{P}}_{\mathrm{emb}},\mathcal{S},\{\mathcal{D}^{(i)}\}_{i=1}^k,E,\epsilon_{\mathrm{suf}})$
\STATE Sample a uniform random permutation $\pi$ on $\{0,\dots,k\}$
\STATE $H\leftarrow [H^{(\pi(0))},H^{(\pi(1))},\dots,H^{(\pi(k))}]$, \quad $j\leftarrow \pi^{-1}(0)$
\IF{$m=\texttt{api-only}$}
    \STATE $\widehat{H}\leftarrow \textsc{ProjectBatch}(H,E,\Pi_{\mathcal V})$
    \STATE $\mathcal{Y}\leftarrow \mathcal{F}_{\mathrm{api}}(\widehat{H})$
\ELSE
    \STATE $\mathcal{Y}\leftarrow \mathcal{F}_{\mathrm{embed}}(H)$
\ENDIF
\STATE $y_{\mathrm{real}}\leftarrow \mathcal{Y}[j]$
\end{algorithmic}
\end{algorithm}
\vspace{-2.0em}
\end{figure}

As shown in Figure~\ref{fig:appendix_workflow} and Algorithm~\ref{alg:dummy_mechanism_new}, given a query $X=[x_1,\dots,x_n]$ and a dummy ratio $r\in[0,1]$, let $T=\lfloor (1-r)n \rfloor$, DP-SPD first splits it into a prefix $\mathcal{P}=[x_1,\dots,x_T]$ and a suffix $\mathcal{S}=[x_{T+1},\dots,x_n]$. The client then perturbs the prefix embeddings with privacy budget $\epsilon_{\mathrm{pre}}$ to preserve prefix alignment and prevent averaging attacks, and then projects the noisy prefix embeddings back to discrete tokens, obtaining a prefix $\widehat{\mathcal{P}}$. The continuation model $G$ uses $\widehat{\mathcal{P}}$ to generate $k$ dummy suffixes:
\[
\mathcal{D}^{(1)},\dots,\mathcal{D}^{(k)} \sim G(\widehat{\mathcal{P}}, n-T).
\]
The candidate suffix set is then
\[
\mathbb{C}=\{\mathcal{S},\mathcal{D}^{(1)},\dots,\mathcal{D}^{(k)}\}.
\]

For $t>T$, the real suffix and each dummy suffix are perturbed independently in \textsc{BuildCandidates}, where $\mathbf{z}_{0,t},\mathbf{z}_{i,t}\sim \mathrm{Lap}(1/\epsilon_{\mathrm{suf}})$. The client then obtains $k+1$ noisy candidates $\{H^{(0)},\dots,H^{(k)}\}$, samples a random permutation $\pi$, and sends the shuffled batch to the server, while recording $j=\pi^{-1}(0)$. Since all candidates share the same prefix, the server can reuse KV caches for that prefix and only compute distinct suffixes. Depending on the deployment mode, the shuffled batch is either sent directly as embeddings to a fully accessible LLM server, or first discretized by nearest-token projection and then sent as token sequences to a black-box API. The auxiliary routines \textsc{PrivatizePrefix}, \textsc{BuildCandidates}, and \textsc{ProjectToTokens} are provided in Appendix~\ref{sec:alg_details}.

\subsection{Privacy Analysis of DP-SPD}
\label{sec:privacy_analysis}

The overall privacy guarantee of DP-SPD consists of two parts: the privacy contribution of the shared prefix and the suffix candidate set, which contains the real suffix together with the dummy suffixes. The shared prefix is locally perturbed, but it does not benefit from suffix-level shuffling amplification. In contrast, the real suffix is perturbed and then hidden among $k$ dummy suffixes through client-side shuffling. We therefore analyze these two components separately.

\begin{theorem}[Privacy Analysis for Prefix]
\label{thm:prefix_privacy}
Let $\mathcal{R}_{\mathrm{pre}}$ be the local randomizer applied to the shared prefix, and suppose that $\mathcal{R}_{\mathrm{pre}}$ satisfies $\epsilon_{\mathrm{pre}}$-$d_\chi$-privacy. The privatized prefix $\widehat{\mathcal{P}}=\mathcal{R}_{\mathrm{pre}}(\mathcal{P})$ satisfies $\epsilon_{\mathrm{pre}}$-$d_\chi$-privacy. Define the exposure of \(\mathcal{P}\) as
\[
\operatorname{Exp}(\mathcal{P})=\{\widehat{\mathcal{P}},\mathcal{D}^{(1)},\ldots,\mathcal{D}^{(k)}\},
\]
where \(D^{(i)} \sim G(\widehat{P}, n-T)\). Then \(\operatorname{Exp}(\mathcal{P})\) also satisfies \(\epsilon_{\mathrm{pre}}\)-\(d_\chi\)-privacy.
\end{theorem}
\begin{proof}
Since $G$ is a randomized post-processing of $\widehat{\mathcal{P}}$, the output of $\mathcal{R}_{\mathrm{pre}}$, the dummy generation step does not directly access the raw prefix $\mathcal{P}$. The exposure
\[
    \mathrm{Exp}(\mathcal{P})
    =
    \bigl\{
    \widehat{\mathcal{P}},
    \mathcal{D}^{(1)},\ldots,\mathcal{D}^{(k)}
    \bigr\},
    \qquad
    \mathcal{D}^{(i)}\sim G(\widehat{\mathcal{P}},n-T),
\]
is therefore a randomized post-processing of $\widehat{\mathcal{P}}=\mathcal{R}_{\mathrm{pre}}(\mathcal{P})$. By the post-processing property of differential privacy~\cite{dwork2014algorithmic}, this transformation cannot increase the privacy loss. Hence, $\mathrm{Exp}(\mathcal{P})$ satisfies the same privacy guarantee as $\mathcal{R}_{\mathrm{pre}}$.
\end{proof}

We next analyze the suffix part. Fix a privatized prefix $\widehat{\mathcal{P}}$. Let $\mathcal{S}$ and $\mathcal{S}'$ be two possible real suffixes under the same prefix, and let $\bm{s}$ and $\bm{s}'$ denote their embedding representations. We define their suffix distance as
\[
c=\|\bm{s}-\bm{s}'\|.
\]
Let $\bm{d}_1,\dots,\bm{d}_k$ denote the dummy suffix embeddings generated from the same privatized prefix $\widehat{\mathcal{P}}$. The two suffix candidate batches are
\[
\mathcal{S}
=
\{\bm{s},\bm{d}_1,\dots,\bm{d}_k\},
\qquad
\mathcal{S}'
=
\{\bm{s}',\bm{d}_1,\dots,\bm{d}_k\}.
\]
Thus, the two batches differ only in the real suffix embedding, while the dummy suffix embeddings follow the same distribution.

\begin{theorem}[Privacy Analysis for Suffix]
\label{thm:suffix_privacy}
Suppose the local randomizer on suffix embeddings satisfies $\epsilon$-$d_\chi$-privacy. Then, for any measurable output set $\mathcal{A}$ and any suffix pair $(\mathcal{S},\mathcal{S}')$ with distance $c=\|\bm{s}-\bm{s}'\|$, DP-SPD satisfies $\epsilon''$-$d_\chi$-privacy with probability at least $1-\delta$, where 

\begin{equation}
    \epsilon'' = \frac{1}{c}\log\left(1+ \frac{e^{c\epsilon}-1}{e^{c\epsilon}+1} \left(\frac{8\sqrt{e^\epsilon\log(4/\delta)}}{\sqrt{k+1}} + \frac{8e^\epsilon}{k+1}\right)\right).
\end{equation}

\end{theorem}

The detailed proof of Theorem~\ref{thm:suffix_privacy} is provided in Appendix~\ref{sec:proofs}. 

\begin{remark}
It is worth emphasizing that despite the difference in privacy definitions, the shuffling amplification under $d_{\chi}$-privacy preserves the same order as in LDP-based analyses. In particular, the effective privacy loss scales as $\mathcal{O}\!\left(1/\sqrt{k+1}\right)$ with respect to the number of dummies. 
\end{remark}

We next establish the privacy guarantee for the complete query.
Under our neighboring-query relation, two queries differ
either in the prefix or in the suffix.

\begin{theorem}[Privacy Guarantee for the Complete Query]
\label{thm:query_privacy}
Let $\mathcal{Q}=(\mathcal{P},\mathcal{S})$ and
$\mathcal{Q}'=(\mathcal{P}',\mathcal{S}')$ be neighboring
queries that differ either in the prefix or in the suffix.
Suppose that the prefix mechanism satisfies
$\epsilon_{\mathrm{pre}}$-$d_\chi$-privacy and that,
for every fixed privatized prefix, the suffix mechanism
satisfies $\epsilon_{\mathrm{suf}}$-$d_\chi$-privacy with probability at least $1-\delta$ in Theorem~\ref{thm:suffix_privacy}.
Then, for any output set $\mathcal{A}$,
\[
\Pr[\mathcal{M}_{\mathrm{DP\text{-}SPD}}(\mathcal{Q})
     \in\mathcal{A}]
\le
e^{\epsilon_{\mathrm{qry}}d_\chi(\mathcal{Q},\mathcal{Q}')}
\Pr[\mathcal{M}_{\mathrm{DP\text{-}SPD}}(\mathcal{Q}')
     \in\mathcal{A}]
+\delta,
\]
where $\epsilon_{\mathrm{qry}}
=
\max\{\epsilon_{\mathrm{pre}},\epsilon_{\mathrm{suf}}\}$.

\end{theorem}

\subsection{Correctness of Answer De-Shuffling}

When the server performs inference on a shuffled batch containing both real and dummy inputs, DP-SPD guarantees correct answer de-shuffling, enabling reliable recovery of the correct answer.

\begin{theorem}[Correctness of Answer De-Shuffling]
Let $f$ be the server-side inference function and let $\pi$ be the client-side permutation. Given the shuffled batch
\[
H=\big[H^{(\pi(0))}, H^{(\pi(1))}, \dots, H^{(\pi(k))}\big],
\]
the client can uniquely recover the output of the real query.
\end{theorem}

\begin{proof}
Since the permutation $\pi$ is generated by the user, the user stores the index $j$ such that $j=\pi^{-1}(0)$.  
After the server returns
\[
\mathcal{Y}=\{f(H^{(\pi(0))}),\dots,f(H^{(\pi(k))})\},
\]
the client retrieves $\mathcal{Y}[j]=f(H^{(0)})$, ensuring correct answer de-shuffling.
\end{proof}

\section{Experiments}

\subsection{Experiment Setups}
\label{sec:setup}

\textbf{Models and Datasets.}
For classification, LLaMA-2-7B-Chat~\cite{touvron2023llama} is evaluated on the IMDb movie review dataset~\cite{maas2011learning}, while \texttt{bert-base-uncased}~\cite{devlin2019bert}  is fine-tuned and evaluated on SST-2 and QNLI from the GLUE benchmark~\cite{wang2018glue}. We report classification accuracy. For generation, we evaluate LLaMA-2-7B-Chat on WikiText-2~\cite{merity2016pointer}, Stanford Alpaca~\cite{taori2023alpaca}, and Databricks-Dolly-15k~\cite{conover2023dolly}. To examine whether DP-SPD generalizes to stronger models and more challenging tasks, we additionally evaluate Qwen3-14B~\cite{yang2025qwen3} on CNN/DailyMail summarization~\cite{see2017get,hermann2015teaching}, reporting perplexity (PPL)~\cite{jelinek1977perplexity} and ROUGE-L~\cite{lin2004rouge}. We further evaluate Qwen3-32B on 1,000 real-world queries from WildChat~\cite{zhao2024wildchat}, comprising 500 answerable QA requests and 500 open-ended requests. We report QA accuracy and open-ended response quality on a 1--5 scale. Dummy suffixes are generated by a lightweight client-side continuation model. We use Qwen2.5-0.5B-Instruct~\cite{qwen2025qwen25technicalreport} for this purpose.

\textbf{Metrics and baselines.}
To directly measure privacy leakage, we additionally evaluate DP-SPD under Query Identification Attack, Embedding Inversion Attack ~\cite{song2020information} and an LLM-assisted attack inspired by the GPT inference attack~\cite{tong2025inferdpt}, where GPT-4o-mini~\cite{openai2024gpt4omini} is used to identify the real candidate and recover the real suffix. We report attack success rate (ASR), real-candidate PPL, dummy-candidate PPL, and the PPL gap between real and dummy candidates for QIA and others.We compare against DP-Forward~\citep{du2023dp}, CAPE~\citep{plant2021cape},
TextObfuscator~\citep{zhou2023textobfuscator}, SANTEXT+~\citep{yue2021differential},
CUSTEXT+~\citep{chen2023customized}, SnD~\citep{mai2024split}, and
InferDPT~\citep{tong2025inferdpt}; incompatible pipelines are implemented
as \emph{style adapters} preserving their core privacy mechanisms.

\subsection{Utility Evaluation}
\label{sec:utility_analysis}

\subsubsection{Classification Task}
\label{sec:classification_utility}

To evaluate whether DP-SPD preserves downstream utility across different
model architectures and task settings, we conduct classification experiments
with BERT-base on SST-2 and QNLI and with LLaMA-2-7B-Chat on IMDb.
Besides comparing against representative privacy-preserving baselines,
we study how the anonymity-set size $k$ and dummy ratio $r$
affect utility under different privacy budgets.
We make the following key observations.

\begin{wrapfigure}[25]{r}{0.423\columnwidth}
\vspace{-2.3\baselineskip}
\centering
\scriptsize
\setlength{\tabcolsep}{3.2pt}
\renewcommand{\arraystretch}{1.05}

\captionsetup{font=footnotesize,skip=3pt}
\captionof{table}{
Classification accuracy (\%) on SST-2 and QNLI using BERT-base at
$\epsilon=3$. “Plaintext” denotes raw-input inference.
}
\label{tab:comparison_sst2}

\resizebox{\linewidth}{!}{
\begin{tabular}{lcc}
\toprule
\textbf{Method} & \textbf{SST-2} & \textbf{QNLI} \\
\midrule
Plaintext & 92.02 & 91.40 \\
DP-Forward~\citep{du2023dp} & 51.69 & 49.48 \\
CAPE~\citep{plant2021cape} & 76.54 & 52.02 \\
TextObf.~\citep{zhou2023textobfuscator} & 79.93 & 58.68 \\
SANTEXT+~\citep{yue2021differential} & 83.71 & 81.64 \\
CUSTEXT+~\citep{chen2023customized} & 79.12 & 77.75 \\
SnD~\citep{mai2024split} & 72.90 & 63.20 \\
\midrule
\textbf{DP-SPD (Embedding)}
& \textbf{91.06}
& \textbf{90.36} \\
\textbf{DP-SPD (API-only)}
& \textbf{90.60}
& \textbf{91.31} \\
\bottomrule
\end{tabular}
}

\vspace{0.2em}

\includegraphics[
    width=0.95\linewidth
]{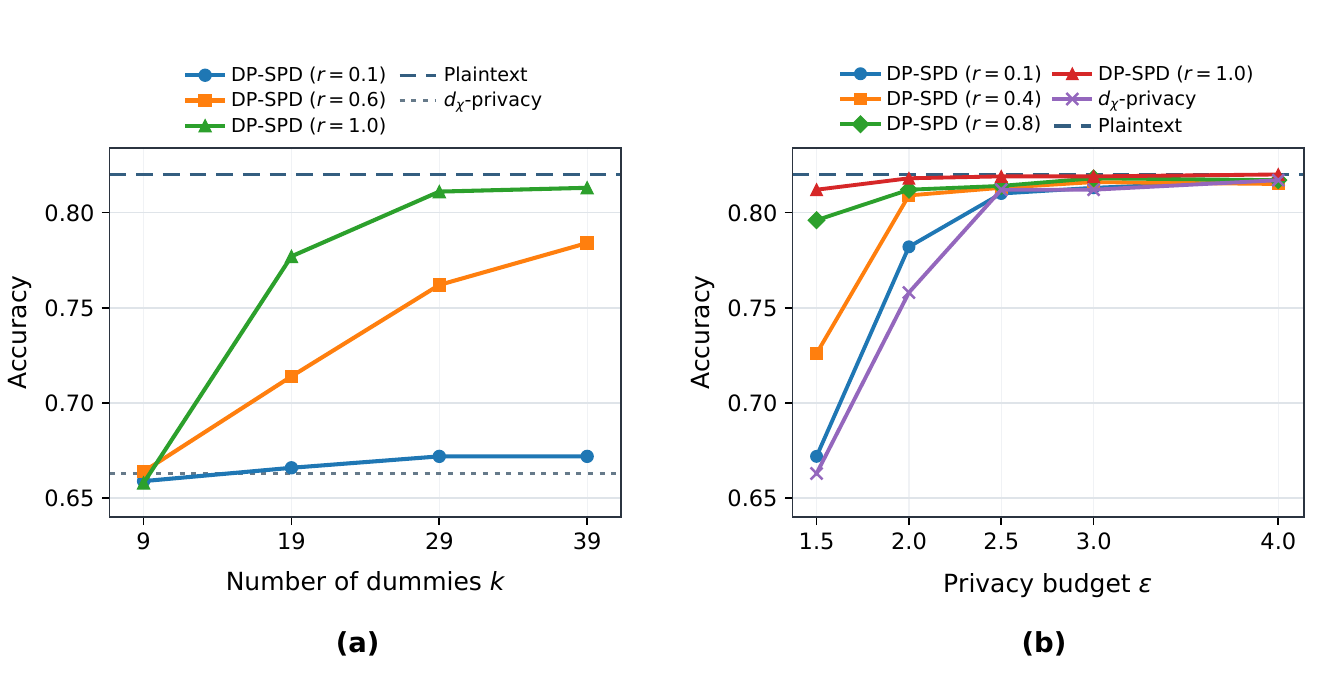}
\captionsetup{font=footnotesize,skip=3pt}
\captionof{figure}{
Classification utility on IMDb using LLaMA-2-7B-Chat.
(a) Accuracy versus $k$ at $\epsilon=1.5$.
(b) Accuracy across privacy budgets with $k=29$.
}
\label{fig:imdb_utility}

\vspace{-1em}
\end{wrapfigure}

\partitle{DP-SPD Preserves Near-Plaintext Classification Utility}
Table~\ref{tab:comparison_sst2} shows that conventional perturbation-based
baselines suffer substantial utility degradation, especially on QNLI:
DP-Forward is close to random guessing, while CAPE, TextObfuscator,
and other token- or representation-perturbation methods remain well below
plaintext inference.
In contrast, both DP-SPD variants retain above 90\% accuracy on SST-2
and QNLI and remain within roughly one percentage point of the plaintext
baseline.
The API-only variant performs comparably to the embedding-access setting,
indicating that nearest-token projection introduces little additional
utility loss.
These results suggest that DP-SPD preserves task-relevant information
more effectively than directly perturbing tokens or intermediate
representations, without requiring an additional denoising model at inference.

\partitle{Larger Anonymity Sets Help Most under Tight Privacy}
Figure~\ref{fig:imdb_utility}(a) shows that increasing $k$ is particularly beneficial when the privacy budget is small.
At $\epsilon=1.5$ and $r=0.6$, increasing $k$ from 9 to 39 raises IMDb
accuracy from 66.4\% to 78.4\%.
The gain also depends strongly on the dummy ratio: at the same
privacy budget, increasing $k$ provides only a small improvement when
$r=0.1$, but a much larger gain for $r=0.6$ and $r=1.0$.
This behavior is consistent with shuffling amplification: when a larger
portion of the query participates in the shuffled candidate set, enlarging
that set allows the mechanism to operate with less perturbation.

\partitle{Protecting a Larger Suffix Further Improves Utility}
Figure~\ref{fig:imdb_utility}(b) further shows that increasing the
dummy ratio $r$ generally improves utility across privacy budgets.
At $\epsilon=1.5$ and $k=29$, accuracy rises from 67.2\% at $r=0.1$
to 81.2\% at $r=1.0$, approaching the 82.0\% plaintext result.
The difference becomes progressively smaller as $\epsilon$ increases,
indicating that the benefit of a larger suffix is most important
in the high-privacy regime.
Taken together, the two ablations show that $k$ and $r$ provide
complementary controls over the privacy--utility trade-off:
larger $k$ strengthens the anonymity set, while larger $r$ allows a larger
fraction of the query to benefit from shuffling amplification.

\subsubsection{Generation Task}
\label{sec:generation_utility}

To evaluate whether DP-SPD preserves utility across different generation
tasks, model scales, and query distributions, we conduct experiments on
WikiText-2, Alpaca, Dolly-15k, CNN/DailyMail, and WildChat using LLaMA-2
and Qwen3 models. We make the following key observations.

\begin{wrapfigure}[17]{r}{0.47\columnwidth}
\vspace{-0.2em}
\centering

\captionsetup{
    font=footnotesize,
    skip=3pt
}

\includegraphics[
    width=\linewidth,
    keepaspectratio
]{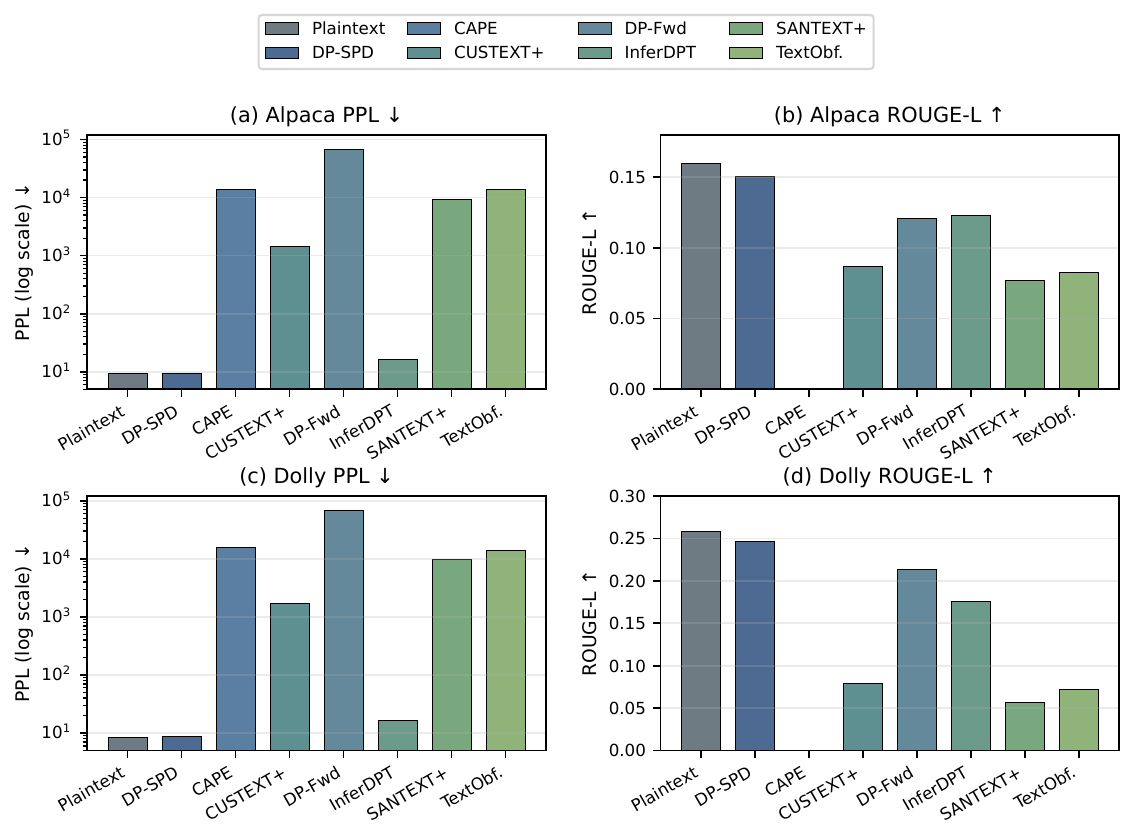}

\caption{
Comparison with related private-inference methods on instruction
generation using LLaMA-2-7B-Chat at $\epsilon=2.5$.
Lower PPL and higher ROUGE-L are better; PPL uses a logarithmic scale.
}
\label{fig:instruction_compare}

\vspace{-0.5\baselineskip}
\end{wrapfigure}

\begin{figure}[t]
    \centering
    \includegraphics[width=0.96\linewidth]{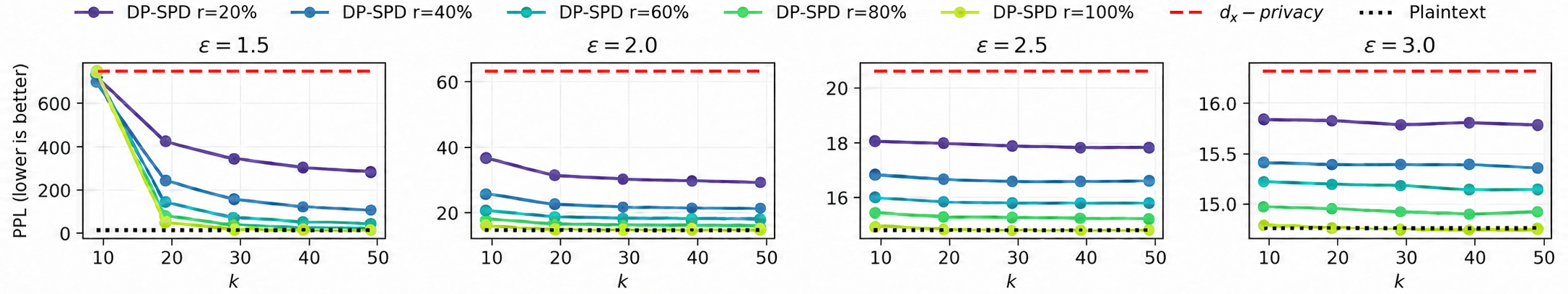}
    \caption{
    Generation utility on WikiText-2 with LLaMA-2-7B-Chat.
    Lower PPL is better.
    We vary the privacy budget $\epsilon$, dummy number $k$, and
    dummy ratio $r$; larger anonymity sets provide the
    largest utility gains under tighter privacy.
    }
    \vspace{-1.5em}
    \label{fig:ppl_main}
\end{figure}

\partitle{DP-SPD Recovers Generation Quality Most Effectively under Tight Privacy}
Figure~\ref{fig:ppl_main} shows that increasing $k$ and $r$ yields the
largest gains on WikiText-2 under small privacy budgets.
With $r=1$ and $k=39$, DP-SPD reaches PPL 14.93 at $\epsilon=2.0$,
close to the plaintext PPL of 14.76; the differences quickly diminish
as $\epsilon$ increases.


\partitle{DP-SPD Preserves Higher Utility than the Evaluated Baselines on Instruction Generation}
As shown in Figure~\ref{fig:instruction_compare}, DP-SPD remains close
to plaintext utility on both Alpaca and Dolly-15k, achieving consistently
lower PPL and higher ROUGE-L than the evaluated style-adapted
private-inference baselines. This advantage holds across both datasets,
indicating that DP-SPD better preserves fluency and task-relevant content
under privacy protection. These results demonstrate its effectiveness
for practical instruction-generation tasks.

\begin{wrapfigure}[11]{r}{0.48\columnwidth}
\vspace{-1.0\baselineskip}
\centering

\scriptsize
\setlength{\tabcolsep}{2.2pt}
\renewcommand{\arraystretch}{1.05}
\captionsetup{
    font=footnotesize,
    skip=4pt
}

\captionof{table}{
Utility on stronger models and real-world queries.
CNN/DailyMail uses Qwen3-14B at $\epsilon=1.5$;
WildChat uses Qwen3-32B at $\epsilon=3$.
}
\label{tab:stronger_realworld}

\resizebox{\linewidth}{!}{
\begin{tabular}{lcc|cc}
\toprule
& \multicolumn{2}{c|}{\textbf{Qwen3-14B / CNN-DM}}
& \multicolumn{2}{c}{\textbf{Qwen3-32B / WildChat}} \\
\cmidrule(lr){2-3}
\cmidrule(lr){4-5}

\textbf{Method}
& \textbf{PPL $\downarrow$}
& \textbf{R-L $\uparrow$}
& \textbf{QA Acc. $\uparrow$}
& \textbf{Open $\uparrow$} \\
\midrule

Plaintext
& 12.92
& 0.1572
& \textbf{72.6\%}
& \textbf{4.716} \\

\textbf{DP-SPD}
& \textbf{14.11}
& \textbf{0.1555}
& \textbf{72.6\%}
& \textbf{4.716} \\

TextObf.
& 16.88
& 0.1499
& 27.6\%
& 2.93 \\

DP-Forward
& 240441.63
& 0.1314
& 21.4\%
& 2.71 \\

CUSTEXT+
& 2606.63
& 0.0352
& 43.8\%
& 3.79 \\

SANTEXT+
& 18081.09
& 0.0500
& 39.6\%
& 3.56 \\

InferDPT
& 36.37
& 0.0831
& 36.6\%
& 3.21 \\

\bottomrule
\end{tabular}
}

\vspace{-0.6\baselineskip}
\end{wrapfigure}

\partitle{The Advantage Persists on Stronger Models and Real-World Queries}
Table~\ref{tab:stronger_realworld} shows that DP-SPD scales to stronger
models and realistic generation settings. On Qwen3-14B/CNN-DailyMail,
it achieves a PPL of 14.11 and ROUGE-L of 0.1555, remaining close to the
plaintext results of 12.92 and 0.1572 while outperforming private
baselines. On Qwen3-32B/WildChat, DP-SPD exactly matches plaintext
utility in QA accuracy and open-ended quality. These results demonstrate
that DP-SPD preserves its utility advantage across model scales,
datasets, and generation formats.

\subsection{Privacy Evaluation}
\label{sec:privacy_eval}

We evaluate DP-SPD against three privacy attacks. Query Identification Attack (QIA) tests whether the real query can be
distinguished from the shuffled candidates using likelihood-based
signals; Embedding Inversion Attack (EIA) first attempts to recover
discrete content from the perturbed representations before identifying
the real candidate; and the LLM-assisted attack uses GPT-4o-mini to
perform semantic candidate selection and query reconstruction.

\begin{wraptable}[9]{r}{0.5\columnwidth}
\vspace{-1.2\baselineskip}
\centering
\scriptsize
\setlength{\tabcolsep}{2.8pt}
\renewcommand{\arraystretch}{1.08}
\captionsetup{
    font=footnotesize,
    skip=3pt
}

\caption{
QIA and dummy-strategy ablation.
Results are averaged over
$\epsilon\in\{1.5,2.0,2.5,3.0\}$,
$r\in\{20\%,50\%,80\%\}$, and
$k\in\{9,19,29\}$.
Lower values are better; bold marks the best ASR and PPL gap.
}
\label{tab:qia_dummy_main}

\resizebox{\linewidth}{!}{
\begin{tabular}{@{}llrrrr@{}}
\toprule
\textbf{Dataset}
& \textbf{Dummy strategy}
& \textbf{ASR (\%)} $\downarrow$
& \textbf{Real PPL}
& \textbf{Dummy PPL}
& \textbf{PPL gap} $\downarrow$ \\
\midrule

\multirow{3}{*}{IMDb}
& DP-SPD prefix-consistent
& \textbf{13.53}
& 29.32
& 36.69
& \textbf{7.38} \\
& Random-sampled
& 36.25
& 29.31
& 49.32
& 20.00 \\
& Random-token
& 97.94
& 29.31
& 1282.39
& 1253.08 \\

\midrule

\multirow{3}{*}{WikiText-2}
& DP-SPD prefix-consistent
& \textbf{11.89}
& 129.61
& 108.92
& \textbf{31.05} \\
& Random-sampled
& 28.83
& 145.49
& 440.03
& 294.54 \\
& Random-token
& 74.86
& 153.44
& 1936.43
& 1782.99 \\

\bottomrule
\end{tabular}
}

\vspace{-0.6\baselineskip}
\end{wraptable}

\partitle{Prefix-Conditioned Dummies Substantially Reduce Query Identification Risk}
Table~\ref{tab:qia_dummy_main} shows that DP-SPD achieves the lowest
QIA ASR and substantially smaller real--dummy PPL gaps on both IMDb
and WikiText-2.
This confirms that conditioning dummy suffixes on the shared prefix
makes them substantially harder to distinguish from the real continuation.

\begin{wraptable}[14]{r}{0.45\columnwidth}
\vspace{-1.2\baselineskip}
\centering
\scriptsize
\renewcommand{\arraystretch}{1.06}
\captionsetup{
    font=scriptsize,
    skip=3pt
}

\captionof{table}{
Group-wise EIA results on IMDb, averaged over
$\epsilon\in\{1.5,3.0\}$ and
$r\in\{20\%,50\%,80\%\}$.
Lower ASR is better.
}
\label{tab:eia_strategy_attack}

\setlength{\tabcolsep}{2.6pt}
\resizebox{\linewidth}{!}{
\begin{tabular}{@{}lccc@{}}
\toprule
\textbf{Dummy strategy}
& \textbf{$k=9$}
& \textbf{$k=19$}
& \textbf{$k=29$} \\
\midrule
DP-SPD prefix-cond.
& \textbf{8.00}
& \textbf{4.50}
& \textbf{2.50} \\
Random-sampled
& 9.50
& 5.33
& 2.67 \\
Random-token
& 8.83
& 5.33
& 3.00 \\
\bottomrule
\end{tabular}
}

\vspace{0.8\baselineskip}

\captionof{table}{
GPT-based attack results.
Lower values indicate stronger privacy protection.
}
\label{tab:gpt_attack}

\setlength{\tabcolsep}{2.4pt}
\resizebox{\linewidth}{!}{
\begin{tabular}{@{}lccc@{}}
\toprule
\textbf{Metric}
& \textbf{$k=9$}
& \textbf{$k=19$}
& \textbf{$k=29$} \\
\midrule
Selection Acc. (\%) $\downarrow$
& 22.67
& 13.33
& \textbf{10.33} \\
Exact Match (\%) $\downarrow$
& 9.25
& 6.83
& \textbf{5.33} \\
Full-Query Token Acc. (\%) $\downarrow$
& 50.09
& \textbf{48.50}
& 48.76 \\
Real-Suffix Token Acc. (\%) $\downarrow$
& 18.35
& 11.41
& \textbf{10.07} \\
\bottomrule
\end{tabular}
}

\vspace{-0.7\baselineskip}
\end{wraptable}

\partitle{DP-SPD Remains Robust under Embedding Inversion}
As shown in Table~\ref{tab:eia_strategy_attack}, increasing $k$ reduces the group-level EIA ASR for all dummy construction strategies. DP-SPD achieves the lowest ASR under both tested dummy numbers. This suggests that DP-SPD also provides stronger candidate-level concealment under embedding inversion.

\partitle{Larger Anonymity Sets Also Suppress Semantic Attacks}
Although GPT-4o-mini is stronger than likelihood-based QIA,
Table~\ref{tab:gpt_attack} shows that increasing $k$ reduces selection
accuracy, exact match, and real-suffix token accuracy. Thus, larger
anonymity sets make candidate identification and private-content
reconstruction difficult for semantic adversaries.

\partitle{API-Only Utility Is Robust to Client Representations}
Changing the client embedding space alters token retention by 20.65
percentage points but GPT-4o accuracy by only 0.8 points. This small
task-level difference indicates that DP-SPD is not strongly tied to a
specific client embedding space and remains effective and reliable in
API-only deployment settings.

\subsection{Efficiency Evaluation}
\label{sec:practicality}

A practical deployment of DP-SPD introduces two additional sources of
overhead: client-side dummy generation and server-side inference over the
augmented candidate set.
We therefore evaluate both the cost of the continuation model and the
benefit of shared-prefix KV-cache reuse.

\begin{wrapfigure}[14]{r}{0.4\linewidth}
\vspace{-1.5em}
\centering
\includegraphics[width=\linewidth]{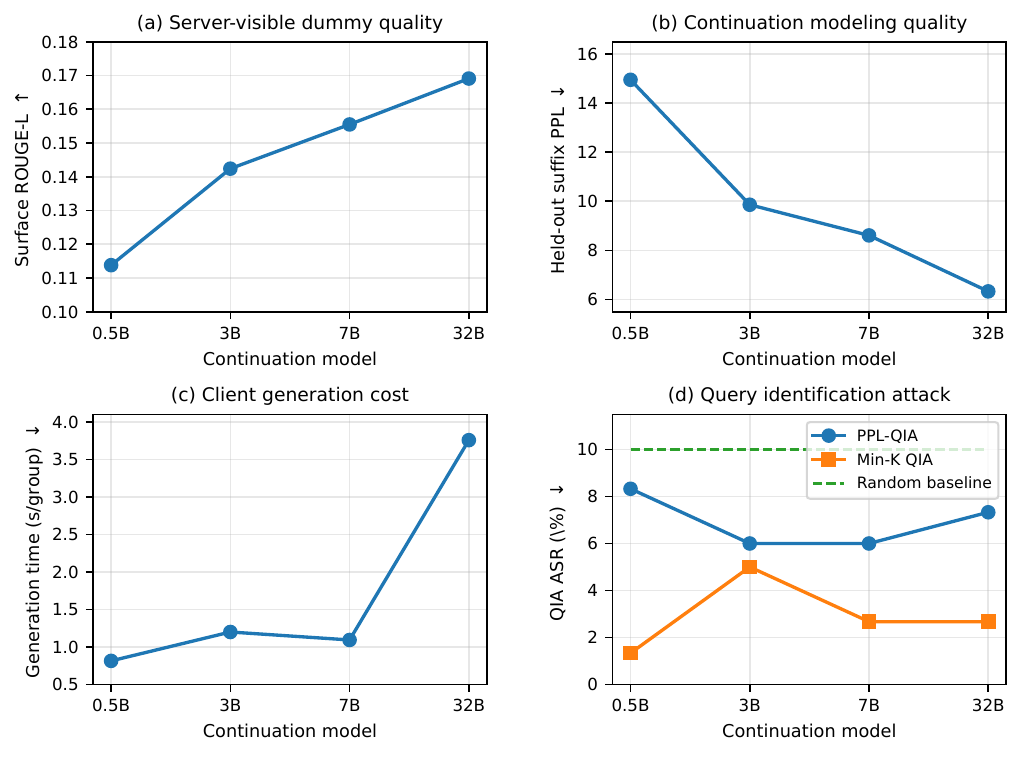}
\captionsetup{
    font=scriptsize,
    skip=3pt
}
\caption{
Effect of continuation-model scale on dummy quality, held-out suffix
PPL, generation time, and QIA ASR.
Higher Surface ROUGE-L is better; lower values are better for all
other metrics.
}
\label{fig:proxy_scale_main}
\end{wrapfigure}

\partitle{Continuation-model scale exposes a clear quality--cost trade-off}
Figure~\ref{fig:proxy_scale_main} studies the client-side continuation
model while keeping the privacy configuration
($\epsilon=3$, $k=9$, $r=0.2$) and evaluation setting fixed.
Larger continuation models generally generate more plausible dummy
suffixes: Surface ROUGE-L increases from 0.1138 with Qwen2.5-0.5B
to 0.1691 with Qwen2.5-32B, while PPL decreases accordingly.
This improvement, however, comes with higher client-side generation
cost; the 32B model requires 3.756\,s per candidate group compared
with 0.813\,s for the 0.5B model.
Importantly, QIA success does not increase monotonically with model
size, suggesting that improving dummy quality does not necessarily make
the real query easier to identify.

\begin{wrapfigure}[16]{r}{0.5\linewidth}
    \centering
    \includegraphics[width=\linewidth]{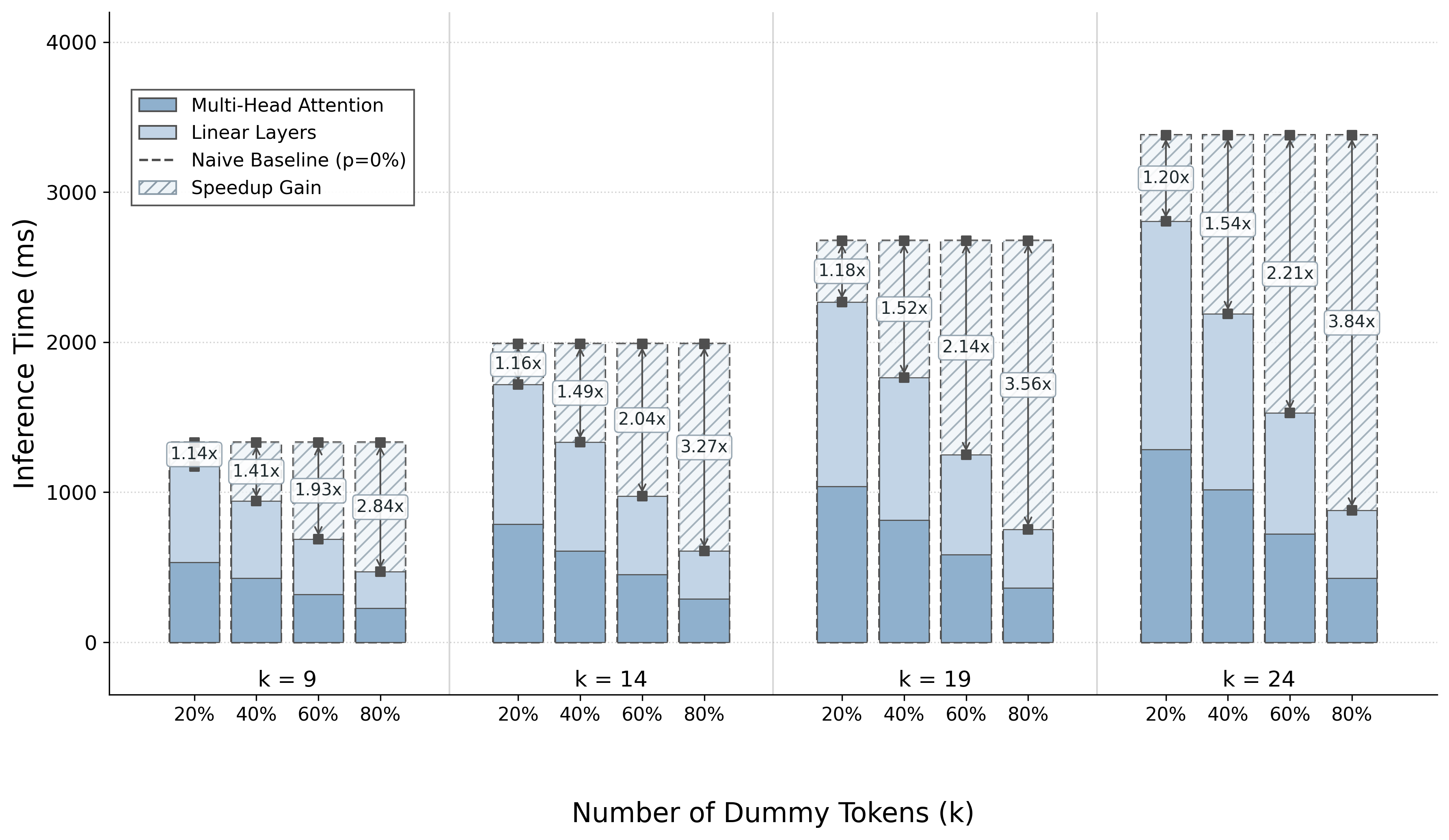}
    \caption{
    Inference efficiency with shared-prefix KV-cache reuse for
    1024-token inputs.
    Larger shared-prefix ratios and larger anonymity sets yield greater
    computational savings.
    }
    \label{fig:efficiency_main}
\end{wrapfigure}

\partitle{Shared-prefix KV caching substantially reduces inference overhead}
Let $p=1-r$ denote the fraction of tokens in the reusable shared prefix.
Without prefix sharing, every real and dummy candidate must be processed independently, increasing inference time with the candidate-set size.
DP-SPD instead computes the shared prefix once and reuses its KV cache across all suffix branches.
As shown in Figure~\ref{fig:efficiency_main}, for 1024-token sequences with $k=24$, shared-prefix reuse improves inference efficiency from 3868.02\,ms to 1002.86\,ms at $p=0.8$, corresponding to a $3.86\times$ speedup.
The benefit becomes more pronounced as the anonymity set grows, with speedup increasing from $2.87\times$ at $k=9$ to $3.86\times$ at $k=24$.

This result also reveals an important practical trade-off.
Increasing the dummy ratio $r$ allows a larger portion of the
query to benefit from dummy-based shuffling, but shortens the reusable
prefix and therefore increases server-side computation.
Conversely, smaller $r$ preserves a longer shared prefix and provides
substantial efficiency gains.
Thus, $r$ and $k$ jointly control not only privacy and utility, but also
inference efficiency, allowing DP-SPD to select practical operating
points without processing every dummy as an independent full sequence.

\section{Conclusion}
We presented DP-SPD, a shuffle-based privacy-preserving inference framework for LLMs. The framework achieves a favorable balance among privacy, utility, and efficiency. Experimental results on classification, generation, and attack-based privacy evaluation show that DP-SPD can substantially reduce inference-time privacy leakage while maintaining utility close to plaintext inference. Our current framework adopts a \textbf{fixed prefix--suffix split}, without explicitly identifying where sensitive information appears in the query. A promising direction is to first detect privacy-sensitive spans and then restructure the query accordingly, for example by moving sensitive content toward positions where shuffle-based protection is more effective, potentially yielding a better privacy--utility trade-off. Another limitation is that our current analysis focuses on single-query inference rather than \textbf{multi-turn conversations} or \textbf{long-context} settings. In such scenarios, repeated occurrences of the same token or semantic content may be perturbed independently into inconsistent outputs, which can accumulate utility degradation across turns or over long contexts. Extending DP-SPD to preserve consistency under repeated queries and long-context composition is therefore an important direction for future work.

\section*{Acknowledgment}
We thank the anonymous reviewers for their insightful comments, which helped improve the quality of this work. We also gratefully acknowledge the support received for this research. This research is supported in part by the National Natural Science Foundation of China (Grant No. 62441238), the National Natural Science Foundation of China under Grant U2441240 (“Ye Qisun” Science Foundation), and the Xiaomi Young Scholars' Key Research Program via Open Competition.

\bibliographystyle{IEEEtran} 
\bibliography{reference}

\clearpage
\appendix

\section{Algorithm Details}
\label{sec:alg_details}

This section provides the detailed procedures used in DP-SPD.
Algorithm~\ref{alg:privatize_prefix} perturbs the prefix embeddings
and projects them back to tokens to obtain the privatized prefix.
Algorithm~\ref{alg:build_candidates} constructs the real and dummy
candidate sequences by reusing the noisy prefix embeddings and
independently perturbing their suffix embeddings.
Algorithm~\ref{alg:projection} specifies the nearest-neighbor
projection used to convert noisy embedding sequences into tokens,
both individually and in batches.

\begin{algorithm}[H]
\caption{\textsc{PrivatizePrefix}}
\label{alg:privatize_prefix}
\begin{algorithmic}[1]
\REQUIRE Prefix $\mathcal{P}=[x_1,\dots,x_T]$, embedding function $E$, projection operator $\Pi_{\mathcal V}$, privacy budget $\epsilon_{\mathrm{pre}}$
\ENSURE Noisy prefix embeddings $\widetilde{\mathcal{P}}_{\mathrm{emb}}$ and privatized token prefix $\widehat{\mathcal{P}}$

\STATE Initialize $\widetilde{\mathcal{P}}_{\mathrm{emb}}\leftarrow [\ ]$
\FOR{$t=1$ to $T$}
    \STATE Sample $\mathbf{z}_t\sim \mathrm{Lap}(1/\epsilon_{\mathrm{pre}})$
    \STATE Append $E(x_t)+\mathbf{z}_t$ to $\widetilde{\mathcal{P}}_{\mathrm{emb}}$
\ENDFOR
\STATE $\widehat{\mathcal{P}}\leftarrow \textsc{ProjectToTokens}(\widetilde{\mathcal{P}}_{\mathrm{emb}},E,\Pi_{\mathcal V})$
\STATE \textbf{return} $(\widetilde{\mathcal{P}}_{\mathrm{emb}},\widehat{\mathcal{P}})$
\end{algorithmic}
\end{algorithm}
\begin{algorithm}[H]
\caption{\textsc{BuildCandidates}}
\label{alg:build_candidates}
\begin{algorithmic}[1]
\REQUIRE Noisy prefix embeddings $\widetilde{\mathcal{P}}_{\mathrm{emb}}$, real suffix $\mathcal{S}$, dummy suffixes $\{\mathcal{D}^{(i)}\}_{i=1}^k$, embedding function $E$, privacy budget $\epsilon_{\mathrm{suf}}$
\ENSURE Noisy candidate sequences $H^{(0)},\dots,H^{(k)}$

\FOR{$i=0$ to $k$}
    \STATE Initialize $H^{(i)}\leftarrow \widetilde{\mathcal{P}}_{\mathrm{emb}}$
\ENDFOR
\FOR{$t=1$ to $|\mathcal{S}|$}
    \STATE Sample $\mathbf{z}_{0,t}\sim\mathrm{Lap}(1/\epsilon_{\mathrm{suf}})$
    \STATE Append $E(\mathcal{S}[t])+\mathbf{z}_{0,t}$ to $H^{(0)}$
    \FOR{$i=1$ to $k$}
        \STATE Sample $\mathbf{z}_{i,t}\sim\mathrm{Lap}(1/\epsilon_{\mathrm{suf}})$
        \STATE Append $E(\mathcal{D}^{(i)}[t])+\mathbf{z}_{i,t}$ to $H^{(i)}$
    \ENDFOR
\ENDFOR
\STATE \textbf{return} $H^{(0)},\dots,H^{(k)}$
\end{algorithmic}
\end{algorithm}
\begin{algorithm}[H]
\caption{\textsc{ProjectToTokens} and \textsc{ProjectBatch}}
\label{alg:projection}
\begin{algorithmic}[1]
\REQUIRE Noisy embedding sequence $Z=[\mathbf{h}_1,\dots,\mathbf{h}_L]$, embedding function $E$, projection operator $\Pi_{\mathcal V}$
\ENSURE Projected token sequence $\widehat{X}$

\STATE $\widehat{X}\leftarrow[\ ]$
\FOR{$\ell=1$ to $L$}
    \STATE $\widehat{x}_{\ell}\leftarrow \Pi_{\mathcal V}(\mathbf{h}_{\ell})=\arg\min_{w\in\mathcal V}\|\mathbf{h}_{\ell}-E(w)\|_2$
    \STATE Append $\widehat{x}_{\ell}$ to $\widehat{X}$
\ENDFOR
\STATE \textbf{return} $\widehat{X}$

\vspace{0.4em}
\STATE \textbf{For a batch:} $\textsc{ProjectBatch}([Z^{(0)},\dots,Z^{(k)}],E,\Pi_{\mathcal V})$
\STATE \hspace{1em} \textbf{return} $[\textsc{ProjectToTokens}(Z^{(0)},E,\Pi_{\mathcal V}),\dots,\textsc{ProjectToTokens}(Z^{(k)},E,\Pi_{\mathcal V})]$
\end{algorithmic}
\end{algorithm}

\section{A Natural-Language Example}
\label{sec:dp_spd_example}

To provide an intuitive illustration of DP-SPD, we present an end-to-end example adapted from a real user query in the WildChat dataset~\cite{zhao2024wildchat}. The example demonstrates query partitioning, prefix-conditioned dummy generation, shuffling, server-side inference, and client-side answer recovery.

\begin{tcolorbox}[
    breakable,
    colback=gray!3,
    colframe=gray!55,
    boxrule=0.5pt,
    arc=1mm,
    left=2mm,
    right=2mm,
    top=1mm,
    bottom=1mm
]
\small

\textbf{Original user query:}

what is a biggest salary for Embedded Developer in Automotive industry with 7 year experience

\vspace{0.5em}
\textbf{Shared prefix $\mathcal{P}$:}

what is a biggest salary for Embedded

\vspace{0.5em}
\textbf{Private suffix $\mathcal{S}$:}

Developer in Automotive industry with 7 year experience

\vspace{0.7em}
\textbf{Shuffled candidates ($k=4$):}

\begin{enumerate}[
    label=\textbf{Candidate \arabic*:},
    start=0,
    leftmargin=*,
    itemsep=3pt,
    topsep=3pt
]
    \item \textit{Dummy}

    what is a biggest salary for Embedded Systems Architect with at least 5 years of experience in Germany

    \item \textit{Real}

    what is a biggest salary for Embedded Developer in Automotive industry with 7 year experience

    \item \textit{Dummy}

    what is a biggest salary for Embedded Systems Engineer in Bangalore, India

    \item \textit{Dummy}

    what is a biggest salary for Embedded Firmware Engineer in China

    \item \textit{Dummy}

    what is a biggest salary for Embedded Systems Programmer with more than 3 years of experience in Europe
\end{enumerate}

\end{tcolorbox}

\noindent\textbf{Privacy perturbation and dummy generation.}
The client first perturbs the shared prefix $\mathcal{P}$ to obtain its privatized representation and the projected prefix $\widehat{\mathcal{P}}$. A lightweight continuation model then generates four dummy suffixes conditioned on $\widehat{\mathcal{P}}$. The real suffix and dummy suffixes are independently perturbed before transmission. For readability, the example displays the underlying natural-language candidates rather than their noisy representations; the projected prefix and suffix tokens may differ after perturbation. The real/dummy labels are provided only for illustration and are not revealed to the server.

\medskip
\noindent\textbf{Shuffling and server-side inference.}
Let $H^{(0)}$ denote the privatized real candidate and $H^{(1)},\ldots,H^{(4)}$ denote the privatized dummy candidates. In this example, the client applies the permutation
\[
\pi=(1,0,2,3,4),
\]
producing the shuffled batch
\[
H=[H^{(1)},H^{(0)},H^{(2)},H^{(3)},H^{(4)}].
\]
The server performs inference on the shuffled, privatized candidates and returns the corresponding responses in the same order. Depending on the deployment mode, the server receives either noisy embeddings or their projected token sequences, rather than the raw queries shown above.

\medskip
\noindent\textbf{Client-side answer de-shuffling.}
The client records the real candidate's position as
\[
j=\pi^{-1}(0)=1.
\]
After receiving the ordered response batch
\[
\mathcal{Y}=[y^{(1)},y^{(0)},y^{(2)},y^{(3)},y^{(4)}],
\]
the client retrieves
\[
y_{\mathrm{real}}=\mathcal{Y}[1]=y^{(0)},
\]
where zero-based indexing is used. Thus, the client recovers the response corresponding to the privatized real query without revealing its position to the server.

\section{Theoretical Proofs}\label{sec:proofs}

\begin{proof}[Proof of Theorem~\ref{thm:suffix_privacy}]
Fix a privatized prefix $\widehat{\mathcal{P}}$. Consider two possible real suffixes under the same prefix, and let $\bm{s}$ and $\bm{s}'$ denote their embedding representations. Define their suffix distance as
\[
c=\|\bm{s}-\bm{s}'\|.
\]

Let $\bm{d}_1,\bm{d}_2,\dots,\bm{d}_k$ be the dummy suffix embeddings generated from the same privatized prefix $\widehat{\mathcal{P}}$. Since the dummy suffix generator uses only $\widehat{\mathcal{P}}$ and is independent of the real suffix, replacing $\bm{s}$ with $\bm{s}'$ changes only the real suffix embedding. We denote the two corresponding suffix candidate groups by
\[
\mathcal{S}
=
\{\bm{s},\bm{d}_1,\bm{d}_2,\dots,\bm{d}_k\},
\qquad
\mathcal{S}'
=
\{\bm{s}',\bm{d}_1,\bm{d}_2,\dots,\bm{d}_k\}.
\]

Following Lemma 3.3 in~\citep{feldman2021hiding}, there exists a randomized post-processing algorithm $f$ such that the canonical shuffled output distributions can be written as
\[
P_0=f(A+1,C-A),
\qquad
Q_0=f(A,C-A+1),
\]
where $C\sim\mathrm{Bin}(k,p)$ and $A\sim\mathrm{Bin}(C,1/2)$. According to Lemma A.1 in~\citep{feldman2021hiding}, if
\[
p\geq\frac{16\ln(2/\delta)}{k+1},
\]
then $P_0$ and $Q_0$ are $(\eta,\delta)$-indistinguishable, where
\[
\eta
=
\log\left(
1+
\frac{8\sqrt{\log(4/\delta)}}{\sqrt{p(k+1)}}
+
\frac{8}{p(k+1)}
\right).
\]

Since the local randomizer on suffix embeddings satisfies $\epsilon$-$d_\chi$-privacy, it satisfies standard pure DP with privacy budget $c\epsilon$ on the pair $(\bm{s},\bm{s}')$. By Lemma 3.4 in~\citep{kairouz2015composition}, there exists a randomized algorithm
\[
\mathcal{Q}:\{0,1\}\rightarrow\mathcal{Y}
\]
such that
\[
\mathcal{R}(\bm{s})
=
\frac{e^{c\epsilon}}{e^{c\epsilon}+1}\mathcal{Q}(0)
+
\frac{1}{e^{c\epsilon}+1}\mathcal{Q}(1),
\]
and
\[
\mathcal{R}(\bm{s}')
=
\frac{1}{e^{c\epsilon}+1}\mathcal{Q}(0)
+
\frac{e^{c\epsilon}}{e^{c\epsilon}+1}\mathcal{Q}(1).
\]

Therefore, the shuffled outputs of the two suffix candidate groups can be expressed as
\[
\mathcal{M}_{DP\text{-}SPD}(\mathcal{S})
\stackrel{d}{=}
P=f(A+\Delta,C-A+1-\Delta),
\]
and
\[
\mathcal{M}_{DP\text{-}SPD}(\mathcal{S}')
\stackrel{d}{=}
Q=f(A+1-\Delta,C-A+\Delta),
\]
where
\[
\Delta
\sim
\mathrm{Bern}\left(
\frac{e^{c\epsilon}}{e^{c\epsilon}+1}
\right).
\]

Let
\[
U=\frac{1}{2}(P_0+Q_0).
\]
Then
\[
P
=
\frac{2}{e^{c\epsilon}+1}U
+
\frac{e^{c\epsilon}-1}
     {e^{c\epsilon}+1}P_0,
\]
and
\[
Q
=
\frac{2}{e^{c\epsilon}+1}U
+
\frac{e^{c\epsilon}-1}
     {e^{c\epsilon}+1}Q_0.
\]

By the hockey-stick divergence argument,
\[
\max\left\{
D_{e^{\eta'}}(U\|P_0),
D_{e^{\eta'}}(Q_0\|U)
\right\}
\leq
D_{e^{\eta'}}(P_0\|Q_0).
\]
Thus, $P$ and $Q$ are $(\eta',p\delta)$-indistinguishable, where
\[
\eta'
\leq
\log\left(
1+
\frac{e^{c\epsilon}-1}
     {e^{c\epsilon}+1}
\left(
\frac{8\sqrt{\log(4/\delta)}}{\sqrt{p(k+1)}}
+
\frac{8}{p(k+1)}
\right)
\right).
\]
Since $p\in(0,1]$, we have $p\delta\leq\delta$. Therefore, for any measurable output set $\mathcal{A}$,
\[
\Pr[\mathcal{M}_{DP\text{-}SPD}(\mathcal{S})\in \mathcal{A}]
\leq
e^{\eta'}
\Pr[\mathcal{M}_{DP\text{-}SPD}(\mathcal{S}')\in \mathcal{A}]
+\delta.
\]

Taking $p=e^{-\epsilon}$ gives
\[
\eta'
\leq
\log\left(
1+
\frac{e^{c\epsilon}-1}
     {e^{c\epsilon}+1}
\left(
\frac{8\sqrt{e^\epsilon\log(4/\delta)}}{\sqrt{k+1}}
+
\frac{8e^\epsilon}{k+1}
\right)
\right).
\]
Writing $\eta'=\epsilon''c$, we obtain
\[
\epsilon''
=
\frac{1}{c}
\log\left(
1+
\frac{e^{c\epsilon}-1}
     {e^{c\epsilon}+1}
\left(
\frac{8\sqrt{e^\epsilon\log(4/\delta)}}{\sqrt{k+1}}
+
\frac{8e^\epsilon}{k+1}
\right)
\right),
\]
which matches the bound stated in Theorem~\ref{thm:suffix_privacy}. Hence, DP-SPD satisfies the stated $\epsilon''$-$d_\chi$ privacy bound for the suffix part with probability at least $1-\delta$.
\end{proof}


\begin{proof}[Proof of Theorem~\ref{thm:query_privacy}]

We consider the two possible cases.

\partitle{Case 1: The queries differ in the prefix.}
In this case, $\mathcal{S}=\mathcal{S}'$.
Thus, $F(\cdot,\mathcal{S})$ is the same randomized
post-processing for both queries.
By Theorem~\ref{thm:prefix_privacy} and the
post-processing property,
\begin{align*}
\Pr[\mathcal{M}_{\mathrm{DP\text{-}SPD}}(\mathcal{Q})
     \in\mathcal{A}]
&\le
e^{\epsilon_{\mathrm{pre}}d_\chi(\mathcal{P},\mathcal{P}')}
\Pr[\mathcal{M}_{\mathrm{DP\text{-}SPD}}(\mathcal{Q}')
     \in\mathcal{A}]\\
&=
e^{\epsilon_{\mathrm{pre}}d_\chi(\mathcal{Q},\mathcal{Q}')}
\Pr[\mathcal{M}_{\mathrm{DP\text{-}SPD}}(\mathcal{Q}')
     \in\mathcal{A}].
\end{align*}

\partitle{Case 2: The queries differ in the suffix.}
In this case, $\mathcal{P}=\mathcal{P}'$, so the
privatized prefix follows the same distribution
$\mu_{\mathcal{P}}$ for both queries.
For every fixed privatized prefix $p$,
Theorem~\ref{thm:suffix_privacy} gives
\[
\Pr[F(p,\mathcal{S})\in\mathcal{A}]
\le
e^{\epsilon_{\mathrm{suf}}d_\chi(\mathcal{S},\mathcal{S}')}
\Pr[F(p,\mathcal{S}')\in\mathcal{A}]
+\delta.
\]
Integrating over $p$ and using
$\int\mu_{\mathcal{P}}(dp)=1$, we obtain
\begin{align*}
\Pr[\mathcal{M}_{\mathrm{DP\text{-}SPD}}(\mathcal{Q})
     \in\mathcal{A}]
&\le
e^{\epsilon_{\mathrm{suf}}d_\chi(\mathcal{S},\mathcal{S}')}
\Pr[\mathcal{M}_{\mathrm{DP\text{-}SPD}}(\mathcal{Q}')
     \in\mathcal{A}]
+\delta\\
&=
e^{\epsilon_{\mathrm{suf}}d_\chi(\mathcal{Q},\mathcal{Q}')}
\Pr[\mathcal{M}_{\mathrm{DP\text{-}SPD}}(\mathcal{Q}')
     \in\mathcal{A}]
+\delta.
\end{align*}

Therefore, both cases satisfy the claimed guarantee
with
$\epsilon_{\mathrm{qry}}
=\max\{\epsilon_{\mathrm{pre}},\epsilon_{\mathrm{suf}}\}$.
\end{proof}


\section{Hardware Configuration}
All experiments were conducted on 4 NVIDIA RTX A6000 GPUs with 48GB memory. Utility experiments with LLaMA-2-7B and Qwen3-14B required approximately 30 and 50 GPU-hours, respectively. Attack-based privacy evaluations required approximately 30 GPU-hours in total. Efficiency experiments were measured on the same hardware with batch size and sequence length specified in Section~\ref{sec:practicality}.

\section{Additional Experimental Results}
\label{sec:additional_experiments}

This appendix provides the complete experimental results that are omitted from the main paper for space. We organize the material following the same three axes as the main text: utility, privacy, and efficiency. Unless otherwise stated, $r$ denotes the dummy ratio and $p=1-r$ denotes the shared-prefix ratio. For methods whose original pipelines are not directly compatible with our LLaMA/Qwen evaluation stack, we report \emph{style-adapted} implementations under a unified model, dataset, and evaluator harness; the exact adaptations are summarized in Table~\ref{tab:adapter_details}.

\subsection{Utility Evaluation Details}
\label{sec:utility_appendix}

\subsubsection{API-only Projection Diagnostics}

Table~\ref{tab:api_only_token_projection} reports the token-level/API-only implementation used in the main classification comparison. The target privacy budget is $\epsilon=3$, with $r=0.6$, $\epsilon_{\mathrm{private}}=3.07$, $\epsilon_{\mathrm{non\mbox{-}sensitive}}=1.8$, clipping norm 1, and seed 42. Special tokens are kept unchanged; all other active tokens are obtained by nearest-neighbor projection of the privatized embedding to the vocabulary of the same checkpoint.

\begin{table}[t]
\centering
\small
\caption{API-only token projection diagnostics at $\epsilon=3$. Token recovery is the fraction of active tokens unchanged after nearest-token projection.}
\label{tab:api_only_token_projection}
\setlength{\tabcolsep}{4pt}
\begin{tabular}{lrrrr}
\toprule
\textbf{Dataset} & \textbf{Samples} & \textbf{Active tokens} & \textbf{Changed} & \textbf{Recovery} \\
\midrule
SST-2 & 872 & 20,199 & 18 & 99.9109\% \\
QNLI  & 5,463 & 262,051 & 323 & 99.8767\% \\
\bottomrule
\end{tabular}
\end{table}

The resulting API-only accuracies are 90.60\% on SST-2 and 91.31\% on QNLI, compared with 91.06\% and 90.36\% for the embedding-level DP-SPD setting. Since each token-level result uses one fixed noise seed, the QNLI difference should be interpreted as run-to-run variation rather than an improvement caused by projection.

To isolate the interface effect, we additionally compare direct DP embedding inference and nearest-token projection on 500 SST-2 examples using five independent noise seeds at $\epsilon=1.5$. The clean embedding and clean token paths are identical, while projection introduces only a 1.67 percentage-point additional loss under DP perturbation (Table~\ref{tab:projection_gap}).

\begin{table}[t]
\centering
\small
\caption{Projection-induced interface gap on SST-2. Results are mean $\pm$ standard deviation over five noise seeds.}
\label{tab:projection_gap}
\begin{tabular}{lcc}
\toprule
\textbf{Input / Interface} & \textbf{Accuracy (\%)} & \textbf{$\Delta$ vs. DP Emb.} \\
\midrule
Clean embedding & 94.20 & -- \\
Clean API-only & 94.20 & -- \\
DP-SPD embedding & $89.13\pm0.24$ & -- \\
DP-SPD API-only projection & $87.46\pm0.31$ & $-1.67$ pp \\
\bottomrule
\end{tabular}
\end{table}

\subsubsection{Classification Results}

We retain the complete IMDb sweeps from the previous version. Table~\ref{tab:cls_results} varies the dummy ratio at $k=29$, while Table~\ref{tab:utility_vs_k} isolates the effect of increasing the anonymity-set size. These results show that the benefit of larger $k$ is strongest under tighter privacy budgets.

\begin{table*}[h]
    \centering
    \caption{Classification Accuracy on IMDb using LLaMA-2-7B ($k=29$).}
    \label{tab:cls_results}
    \begin{tabular}{c|c|ccccc}
        \toprule
        \multirow{2}{*}{\textbf{Method}} & \multirow{2}{*}{\textbf{Dummy Ratio}} & \multicolumn{5}{c}{\textbf{Privacy Budget ($\epsilon$)}} \\
         & & \textbf{1.5} & \textbf{2.0}  & \textbf{2.5} & \textbf{3.0} & \textbf{4.0} \\
        \midrule
        \textbf{Plaintext} & - & \multicolumn{5}{c}{0.820} \\
        \midrule
        $d_{\chi}$-privacy & - & 0.663 & 0.758 & 0.812 & 0.812 & 0.817 \\
        \midrule
        \multirow{5}{*}{DP-SPD} 
         & $r=0.1$ & 0.672 & 0.782 & 0.810 & 0.813 & 0.817 \\
         & $r=0.2$ & 0.689 & 0.797 & 0.809 & 0.813 & 0.815 \\
         & $r=0.4$ & 0.726 & 0.809 & 0.813 & 0.816 & 0.815 \\
         & $r=0.6$ & 0.762 & 0.811 & 0.813 & 0.817 & 0.816 \\
         & $r=0.8$ & 0.796 & 0.812 & 0.814 & 0.818 & 0.817 \\
         & $r=1.0$ & 0.812 & 0.818 & 0.819 & 0.819 & 0.820 \\
        \bottomrule
    \end{tabular}
\end{table*}

\begin{table}[h]
    \centering
    \caption{The impact of the number of dummy inputs ($k$) on LLaMA-2-7B model accuracy under different privacy budgets ($\epsilon$) using IMDb dataset. The results demonstrate that utility generally improves as $k$ increases.}
    \label{tab:utility_vs_k}
    \small
    \setlength{\tabcolsep}{3pt} 
    \renewcommand{\arraystretch}{1.1}
    \begin{tabular}{l c c c c c c}
        \toprule
        \textbf{Method} & \textbf{Dummy Ratio} & \textbf{$\epsilon$} & \textbf{$k=9$} & \textbf{$k=19$} & \textbf{$k=29$} & \textbf{$k=39$} \\
        \midrule
        Plaintext & - & - & \multicolumn{4}{c}{0.820} \\
        \midrule
        \multirow{3}{*}{DP-SPD} & \multirow{3}{*}{0.1}
          & 1.5 & 0.659 & 0.666 & 0.672 & 0.672 \\
        & & 1.8 & 0.724 & 0.732 & 0.737 & 0.741 \\
        & & 2.0 & 0.779 & 0.781 & 0.782 & 0.782 \\
        \midrule
        \multirow{3}{*}{DP-SPD} & \multirow{3}{*}{0.6}
          & 1.5 & 0.664 & 0.714 & 0.762 & 0.784 \\
        & & 1.8 & 0.788 & 0.806 & 0.808 & 0.807 \\
        & & 2.0 & 0.810 & 0.812 & 0.811 & 0.814 \\
        \midrule
        \multirow{3}{*}{DP-SPD} & \multirow{3}{*}{1.0}
          & 1.5 & 0.658 & 0.777 & 0.811 & 0.813 \\
        & & 1.8 & 0.810 & 0.816 & 0.814 & 0.815 \\
        & & 2.0 & 0.814 & 0.816 & 0.818 & 0.816 \\
        \bottomrule
    \end{tabular}%
\end{table}

\begin{table*}[t]
\centering
\small
\caption{IMDb classification accuracy (\%) under the unified LLaMA-2-7B evaluation harness. DP-SPD uses the fixed configuration $r=1,k=29$; the remaining private baselines are style-adapted implementations.}
\label{tab:imdb_adapter_sweep}
\setlength{\tabcolsep}{5pt}
\begin{tabular}{lccccc}
\toprule
\textbf{Method} & $\boldsymbol{\epsilon=1.5}$ & $\boldsymbol{\epsilon=2.0}$ & $\boldsymbol{\epsilon=2.5}$ & $\boldsymbol{\epsilon=3.0}$ & $\boldsymbol{\epsilon=4.0}$ \\
\midrule
Clean control & 81.85 & 81.85 & 81.85 & 81.85 & 81.85 \\
\textbf{DP-SPD} ($r=1,k=29$) & \textbf{81.20} & \textbf{81.80} & \textbf{81.90} & \textbf{81.90} & \textbf{82.00} \\
CAPE-style & 50.98 & 50.68 & 50.92 & 51.17 & 51.72 \\
CUSTEXT-style & 59.62 & 59.38 & 59.68 & 60.68 & 63.45 \\
DP-Forward-style & 50.28 & 50.30 & 50.13 & 50.16 & 50.28 \\
InferDPT-style & 50.40 & 50.49 & 52.43 & 52.88 & 54.46 \\
SANTEXT-style & 50.39 & 50.66 & 50.38 & 50.49 & 50.49 \\
TextObfuscator-style & 49.96 & 50.27 & 51.18 & 53.32 & 56.27 \\
\bottomrule
\end{tabular}
\end{table*}

\begin{figure}[t]
    \centering
    \includegraphics[width=0.92\linewidth]{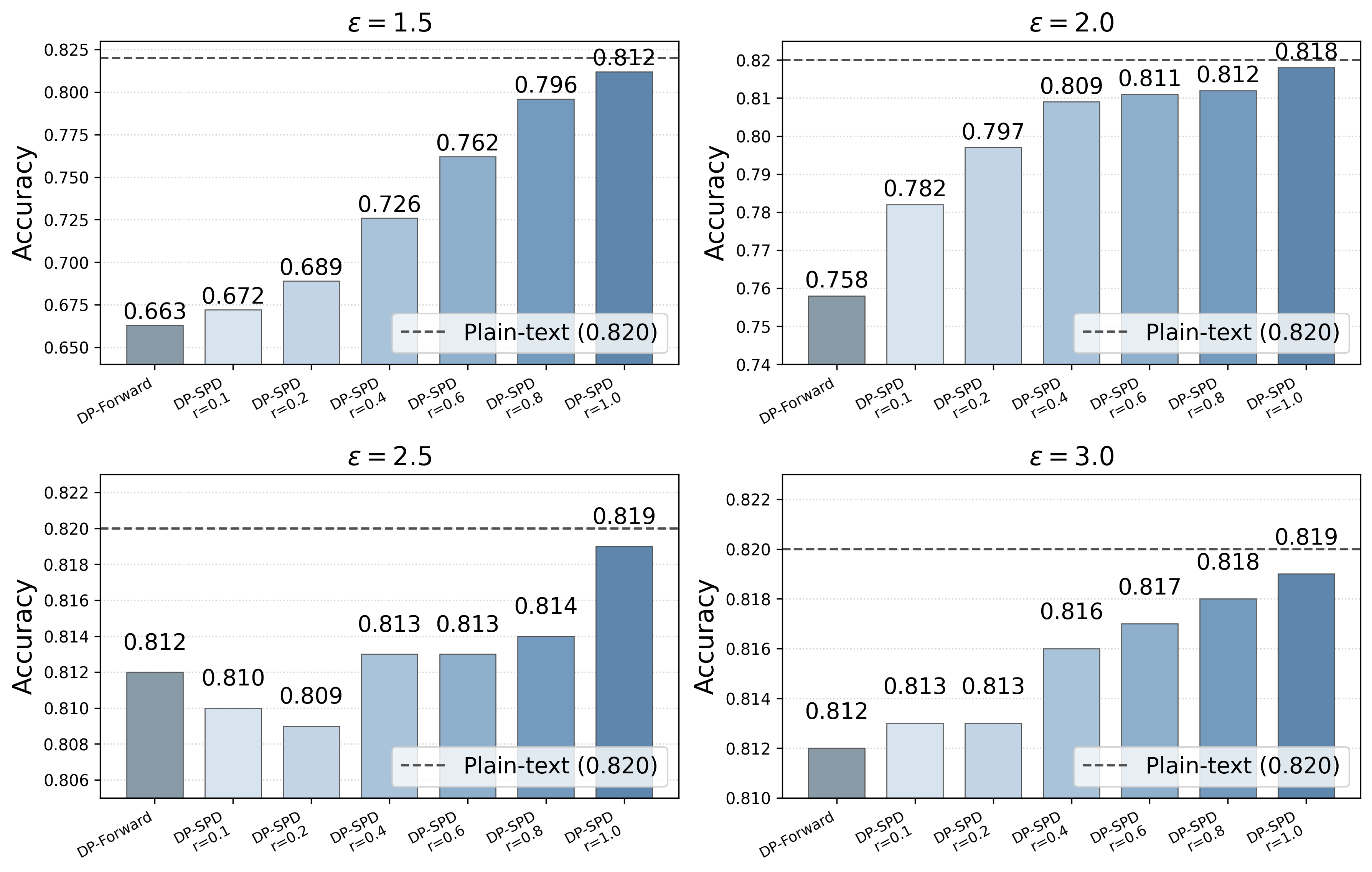}
    \caption{IMDb classification accuracy under different privacy budgets and dummy ratios. This visualization was moved from the main text to the appendix for space.}
    \label{fig:imdb_appendix}
\end{figure}

\subsubsection{Generation Results on WikiText-2, Alpaca, and Dolly}

The following tables retain the complete DP-SPD parameter sweeps from the previous version and add unified style-adapter comparisons. For the cross-method comparison, DP-SPD is evaluated at a fixed configuration for every privacy budget rather than selecting the best $r$ or $k$ separately at each $\epsilon$.

\begin{table*}[h]
    \centering
    \caption{Utility comparison (PPL and ROUGE-L) on \textbf{WikiText-2}. We compare Baselines and DP-SPD with varying Dummy Ratios ($r$). 
    At $\epsilon=1.5$, utility is generally poor due to extreme noise.
    At $\epsilon=2.0$, increasing $r$ significantly improves utility by reducing the impact of context noise. 
    At $\epsilon \ge 2.5$, DP-SPD converges to the plaintext baseline regardless of $r$.}
    \label{tab:utility_results}
    \resizebox{\linewidth}{!}{
    \begin{tabular}{c|c|ccccc|ccccc|ccccc|ccccc}
        \toprule
        \multirow{2}{*}{\textbf{Method}} &
        \multirow{2}{*}{\textbf{Metric}} &
        \multicolumn{5}{c|}{$\boldsymbol{\epsilon = 1.5}$} &
        \multicolumn{5}{c|}{$\boldsymbol{\epsilon = 2.0}$} &
        \multicolumn{5}{c|}{$\boldsymbol{\epsilon = 2.5}$} &
        \multicolumn{5}{c}{$\boldsymbol{\epsilon = 3.0}$} \\
        & 
        & $k=9$ & $k=19$ & $k=29$ & $k=39$ & $k=49$
        & $k=9$ & $k=19$ & $k=29$ & $k=39$ & $k=49$
        & $k=9$ & $k=19$ & $k=29$ & $k=39$ & $k=49$
        & $k=9$ & $k=19$ & $k=29$ & $k=39$ & $k=49$ \\
        \midrule
        \textbf{Plaintext}
        & PPL & \multicolumn{20}{c}{14.76} \\
        & ROUGE & \multicolumn{20}{c}{0.194} \\
        \midrule
        \multirow{2}{*}{$d_{\chi}$-privacy}
        & PPL
        & \multicolumn{5}{c|}{747.61}
        & \multicolumn{5}{c|}{63.14}
        & \multicolumn{5}{c|}{20.59}
        & \multicolumn{5}{c}{16.31} \\
        & ROUGE
        & \multicolumn{5}{c|}{0.037}
        & \multicolumn{5}{c|}{0.129}
        & \multicolumn{5}{c|}{0.174}
        & \multicolumn{5}{c}{0.193} \\
        \midrule
        \multicolumn{22}{c}{\textit{DP-SPD with varying Dummy Ratio } ($r$)} \\
        \midrule
        \multirow{2}{*}{$r=0.1$}
        & PPL
        & 734.12 & 570.66 & 525.02 & 490.48 & 484.47
        & 48.00 & 42.64 & 41.58 & 41.12 & 40.59
        & 19.01 & 18.92 & 18.91 & 18.89 & 18.85
        & 16.09 & 16.07 & 16.06 & 16.07 & 16.07 \\
        & ROUGE
        & 0.040 & 0.044 & 0.050 & 0.054 & 0.054
        & 0.138 & 0.146 & 0.147 & 0.142 & 0.146
        & 0.181 & 0.181 & 0.183 & 0.185 & 0.181
        & 0.190 & 0.192 & 0.192 & 0.192 & 0.191 \\
        \cmidrule{1-22}
        \multirow{2}{*}{$r=0.2$}
        & PPL
        & 719.85 & 423.16 & 344.25 & 305.93 & 284.33
        & 36.76 & 31.58 & 30.33 & 29.88 & 29.44
        & 18.03 & 17.95 & 17.85 & 17.80 & 17.80
        & 15.83 & 15.82 & 15.78 & 15.80 & 15.78 \\
        & ROUGE
        & 0.037 & 0.054 & 0.069 & 0.072 & 0.079
        & 0.154 & 0.153 & 0.157 & 0.153 & 0.155
        & 0.180 & 0.185 & 0.184 & 0.182 & 0.180
        & 0.187 & 0.189 & 0.185 & 0.186 & 0.182 \\
        \cmidrule{1-22}
        \multirow{2}{*}{$r=0.4$}
        & PPL
        & 692.66 & 244.47 & 156.92 & 122.27 & 106.90
        & 25.79 & 22.59 & 21.89 & 21.56 & 21.41
        & 16.79 & 16.62 & 16.54 & 16.55 & 16.57
        & 15.41 & 15.39 & 15.39 & 15.39 & 15.36 \\
        & ROUGE
        & 0.044 & 0.075 & 0.098 & 0.115 & 0.119
        & 0.167 & 0.174 & 0.174 & 0.177 & 0.179
        & 0.186 & 0.188 & 0.183 & 0.185 & 0.189
        & 0.190 & 0.194 & 0.194 & 0.196 & 0.193 \\
        \cmidrule{1-22}
        \multirow{2}{*}{$r=0.6$}
        & PPL
        & 740.58 & 144.90 & 72.82 & 52.68 & 45.51
        & 20.85 & 18.70 & 18.29 & 18.20 & 18.02
        & 15.95 & 15.81 & 15.76 & 15.76 & 15.77
        & 15.22 & 15.19 & 15.18 & 15.14 & 15.14 \\
        & ROUGE
        & 0.043 & 0.096 & 0.131 & 0.142 & 0.143
        & 0.176 & 0.181 & 0.186 & 0.182 & 0.187
        & 0.191 & 0.192 & 0.189 & 0.192 & 0.191
        & 0.195 & 0.192 & 0.193 & 0.200 & 0.196 \\
        \cmidrule{1-22}
        \multirow{2}{*}{$r=0.8$}
        & PPL
        & 735.70 & 82.00 & 36.53 & 27.42 & 24.51
        & 17.95 & 16.52 & 16.31 & 16.20 & 16.09
        & 15.40 & 15.23 & 15.22 & 15.18 & 15.16
        & 14.97 & 14.95 & 14.92 & 14.90 & 14.92 \\
        & ROUGE
        & 0.039 & 0.115 & 0.150 & 0.167 & 0.170
        & 0.185 & 0.184 & 0.192 & 0.189 & 0.190
        & 0.193 & 0.188 & 0.193 & 0.193 & 0.193
        & 0.195 & 0.192 & 0.195 & 0.199 & 0.195 \\
        \midrule
        \multirow{2}{*}{$r=1.0$}
        & PPL
        & 745.59 & 48.96 & 21.52 & 17.54 & 16.47
        & 16.23 & 15.21 & 15.01 & 14.93 & 14.89
        & 14.92 & 14.82 & 14.79 & 14.78 & 14.77
        & 14.79 & 14.76 & 14.75 & 14.74 & 14.75 \\
        & ROUGE
        & 0.036 & 0.133 & 0.172 & 0.190 & 0.196
        & 0.194 & 0.193 & 0.194 & 0.192 & 0.194
        & 0.189 & 0.196 & 0.196 & 0.199 & 0.195
        & 0.198 & 0.198 & 0.194 & 0.196 & 0.193 \\
        \bottomrule
    \end{tabular}
    }
\end{table*}

\begin{table*}[t]
\centering
\small
\caption{WikiText-2 generation comparison under the unified LLaMA-2-7B harness. Each entry is PPL / ROUGE-L. DP-SPD uses $r=1,k=39$.}
\label{tab:wikitext_adapter_compare}
\setlength{\tabcolsep}{4.5pt}
\begin{tabular}{lcccc}
\toprule
\textbf{Method} & $\boldsymbol{\epsilon=1.5}$ & $\boldsymbol{\epsilon=2.0}$ & $\boldsymbol{\epsilon=2.5}$ & $\boldsymbol{\epsilon=3.0}$ \\
\midrule
Plaintext & 14.76 / .194 & 14.76 / .194 & 14.76 / .194 & 14.76 / .194 \\
\textbf{DP-SPD} & \textbf{17.54 / .190} & \textbf{14.93 / .192} & \textbf{14.78 / .199} & \textbf{14.74 / .196} \\
CAPE-style & 21853.10 / .000 & 20549.84 / .000 & 19633.91 / .000 & 18942.00 / .000 \\
CUSTEXT-style & 3177.39 / .095 & 2938.88 / .093 & 2647.31 / .097 & 2297.97 / .100 \\
DP-Forward-style & 87500.65 / .168 & 78406.94 / .168 & 68875.98 / .163 & 60178.53 / .158 \\
SANTEXT-style & 10729.42 / .090 & 10697.31 / .089 & 10623.72 / .084 & 10452.40 / .085 \\
TextObfuscator-style & 34437.38 / .090 & 29455.69 / .087 & 14533.64 / .092 & 7473.66 / .095 \\
\bottomrule
\end{tabular}
\end{table*}

For reference, the InferDPT-style extractor-side conditional PPLs are 77.32, 77.00, 82.12, and 81.69 for $\epsilon=1.5,2.0,2.5,3.0$, respectively; these values are reported only as diagnostics and are not used for cross-method PPL ranking.

\begin{table*}[h]
    \centering
    \caption{Utility comparison (PPL and ROUGE-L) on the cleaned version of \textbf{Stanford Alpaca}. We compare Baselines and DP-SPD with varying Dummy Ratios ($r$).
    Consistent with WikiText-2 results, utility is low at $\epsilon=1.5$ but improves significantly with higher $r$ and larger $\epsilon$.
    DP-SPD shows clear benefits from context preservation at lower $\epsilon$.}
    \label{tab:utility_results_alpaca}
    \resizebox{\linewidth}{!}{
    \begin{tabular}{c|c|cccc|cccc|cccc}
        \toprule
        \multirow{2}{*}{\textbf{Method}} &
        \multirow{2}{*}{\textbf{Metric}} &
        \multicolumn{4}{c|}{$\boldsymbol{\epsilon = 1.5}$} &
        \multicolumn{4}{c|}{$\boldsymbol{\epsilon = 2.5}$} &
        \multicolumn{4}{c}{$\boldsymbol{\epsilon = 4.0}$} \\
        & 
        & $k=9$ & $k=19$ & $k=29$ & $k=39$
        & $k=9$ & $k=19$ & $k=29$ & $k=39$
        & $k=9$ & $k=19$ & $k=29$ & $k=39$ \\
        \midrule
        \textbf{Plaintext}
        & PPL & \multicolumn{12}{c}{9.34} \\
        & ROUGE & \multicolumn{12}{c}{0.160} \\
        \midrule
        \multirow{2}{*}{$d_{\chi}$-privacy}
        & PPL
        & \multicolumn{4}{c|}{431.93}
        & \multicolumn{4}{c|}{14.43}
        & \multicolumn{4}{c}{9.56} \\
        & ROUGE
        & \multicolumn{4}{c|}{0.059}
        & \multicolumn{4}{c|}{0.134}
        & \multicolumn{4}{c}{0.144} \\
        \midrule
        \multicolumn{14}{c}{\textit{DP-SPD with varying Dummy Ratio } ($r$)} \\
        \midrule
        \multirow{2}{*}{$r=0.2$}
        & PPL
        & 433.55 & 256.42 & 211.78 & 191.67
        & 11.90 & 11.60 & 11.55 & 11.49
        & 9.48 & 9.53 & 9.48 & 9.51 \\
        & ROUGE
        & 0.050 & 0.071 & 0.072 & 0.075
        & 0.131 & 0.130 & 0.137 & 0.133
        & 0.145 & 0.145 & 0.149 & 0.147 \\
        \cmidrule{1-14}
        \multirow{2}{*}{$r=0.4$}
        & PPL
        & 446.27 & 156.40 & 106.99 & 83.95
        & 10.86 & 10.55 & 10.45 & 10.43
        & 9.43 & 9.44 & 9.45 & 9.45 \\
        & ROUGE
        & 0.047 & 0.094 & 0.100 & 0.109
        & 0.137 & 0.136 & 0.136 & 0.144
        & 0.146 & 0.147 & 0.150 & 0.149 \\
        \midrule
        \multirow{2}{*}{$r=1.0$}
        & PPL
        & 427.13 & 34.28 & 15.06 & 11.87
        & 9.37 & 9.24 & 9.25 & 9.25
        & 9.47 & 9.48 & 9.48 & 9.48 \\
        & ROUGE
        & 0.057 & 0.128 & 0.133 & 0.132
        & 0.144 & 0.144 & 0.145 & 0.151
        & 0.155 & 0.151 & 0.152 & 0.153 \\
        \bottomrule
    \end{tabular}
    }
\end{table*}

\begin{table*}[t!]
    \centering
    \caption{Utility comparison (PPL and ROUGE-L) on \textbf{Databricks-Dolly-15k}.
    The trends are consistent with other datasets: DP-SPD effectively balances privacy and utility, particularly at moderate noise levels ($\epsilon=2.5$).
    At high privacy ($\epsilon=1.5$), DP-SPD with higher $r$ shows improved text quality (ROUGE).}
    \label{tab:utility_results_dolly}
    \resizebox{\linewidth}{!}{
    \begin{tabular}{c|c|cccc|cccc|cccc}
        \toprule
        \multirow{2}{*}{\textbf{Method}} &
        \multirow{2}{*}{\textbf{Metric}} &
        \multicolumn{4}{c|}{$\boldsymbol{\epsilon = 1.5}$} &
        \multicolumn{4}{c|}{$\boldsymbol{\epsilon = 2.5}$} &
        \multicolumn{4}{c}{$\boldsymbol{\epsilon = 4.0}$} \\
        & 
        & $k=9$ & $k=19$ & $k=29$ & $k=39$
        & $k=9$ & $k=19$ & $k=29$ & $k=39$
        & $k=9$ & $k=19$ & $k=29$ & $k=39$ \\
        \midrule
        \textbf{Plaintext}
        & PPL & \multicolumn{12}{c}{8.55} \\
        & ROUGE & \multicolumn{12}{c}{0.258} \\
        \midrule
        \multirow{2}{*}{$d_{\chi}$-privacy}
        & PPL
        & \multicolumn{4}{c|}{409.17}
        & \multicolumn{4}{c|}{13.62}
        & \multicolumn{4}{c}{9.31} \\
        & ROUGE
        & \multicolumn{4}{c|}{0.064}
        & \multicolumn{4}{c|}{0.234}
        & \multicolumn{4}{c}{0.243} \\
        \midrule
        \multicolumn{14}{c}{\textit{DP-SPD with varying Dummy Ratio } ($r$)} \\
        \midrule
        \multirow{2}{*}{$r=0.2$}
        & PPL
        & 382.84 & 228.19 & 178.00 & 160.03
        & 11.41 & 11.24 & 11.18 & 11.07
        & 9.11 & 9.06 & 9.05 & 9.06 \\
        & ROUGE
        & 0.063 & 0.093 & 0.116 & 0.114
        & 0.234 & 0.245 & 0.244 & 0.250
        & 0.240 & 0.248 & 0.246 & 0.243 \\
        \cmidrule{1-14}
        \multirow{2}{*}{$r=0.4$}
        & PPL
        & 393.74 & 144.08 & 89.58 & 65.81
        & 10.47 & 10.26 & 10.23 & 10.17
        & 8.89 & 8.92 & 8.88 & 8.89 \\
        & ROUGE
        & 0.075 & 0.096 & 0.133 & 0.161
        & 0.241 & 0.237 & 0.239 & 0.236
        & 0.257 & 0.249 & 0.252 & 0.246 \\
        \midrule
        \multirow{2}{*}{$r=1.0$}
        & PPL
        & 410.38 & 33.29 & 14.34 & 11.31
        & 9.10 & 8.86 & 8.80 & 8.73
        & 8.61 & 8.63 & 8.63 & 8.61 \\
        & ROUGE
        & 0.062 & 0.181 & 0.233 & 0.240
        & 0.241 & 0.253 & 0.245 & 0.247
        & 0.259 & 0.253 & 0.249 & 0.253 \\
        \bottomrule
    \end{tabular}
    }
\end{table*}

\begin{table*}[t]
\centering
\scriptsize
\caption{Instruction-generation comparison under the unified LLaMA-2-7B harness. Each entry is PPL / ROUGE-L. DP-SPD uses $r=1,k=39$.}
\label{tab:instruction_adapter_compare}
\setlength{\tabcolsep}{3.5pt}
\begin{tabular}{llccc}
\toprule
\textbf{Dataset} & \textbf{Method} & $\boldsymbol{\epsilon=1.5}$ & $\boldsymbol{\epsilon=2.5}$ & $\boldsymbol{\epsilon=4.0}$ \\
\midrule
\multirow{8}{*}{Alpaca}
& Plaintext & 9.34 / .160 & 9.34 / .160 & 9.34 / .160 \\
& \textbf{DP-SPD} & \textbf{11.87 / .132} & \textbf{9.25 / .151} & \textbf{9.48 / .153} \\
& CAPE-style & 17191.15 / .000 & 14053.85 / .000 & 12754.43 / .000 \\
& CUSTEXT-style & 1856.30 / .085 & 1445.68 / .087 & 832.20 / .092 \\
& DP-Forward-style & 86864.22 / .116 & 67077.54 / .121 & 48869.19 / .121 \\
& SANTEXT-style & 9374.88 / .078 & 9253.24 / .077 & 9031.51 / .076 \\
& TextObfuscator-style & 30357.85 / .082 & 13963.50 / .083 & 2732.50 / .093 \\
\midrule
\multirow{8}{*}{Dolly}
& Plaintext & 8.55 / .258 & 8.55 / .258 & 8.55 / .258 \\
& \textbf{DP-SPD} & \textbf{11.31 / .240} & \textbf{8.73 / .247} & \textbf{8.61 / .253} \\
& CAPE-style & 18498.71 / .000 & 15722.25 / .000 & 14218.52 / .000 \\
& CUSTEXT-style & 2050.00 / .080 & 1679.10 / .079 & 950.35 / .091 \\
& DP-Forward-style & 88334.92 / .219 & 69046.99 / .214 & 49849.64 / .215 \\
& SANTEXT-style & 9796.14 / .062 & 9652.38 / .057 & 9482.65 / .060 \\
& TextObfuscator-style & 30970.81 / .075 & 13930.29 / .072 & 2953.12 / .070 \\
\bottomrule
\end{tabular}
\end{table*}

The corresponding InferDPT-style extractor conditional PPLs are 15.89/16.60/16.94 on Alpaca and 16.29/16.62/16.50 on Dolly for $\epsilon=1.5/2.5/4.0$.

\begin{figure*}[t]
    \centering
    \begin{subfigure}{0.47\linewidth}
        \centering
        \includegraphics[width=\linewidth]{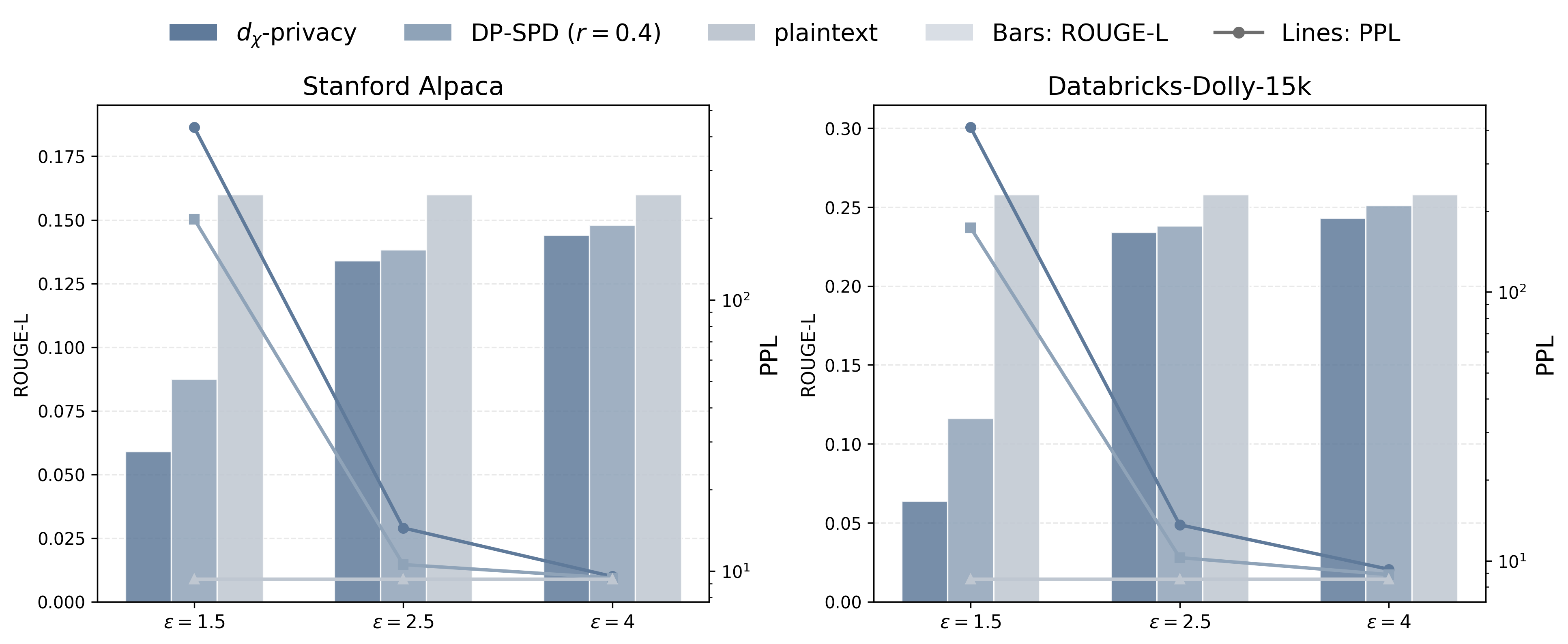}
        \caption{LLaMA-2-7B on Alpaca and Dolly-15k.}
        \label{fig:gen_appendix}
    \end{subfigure}
    \hfill
    \begin{subfigure}{0.47\linewidth}
        \centering
        \includegraphics[width=\linewidth]{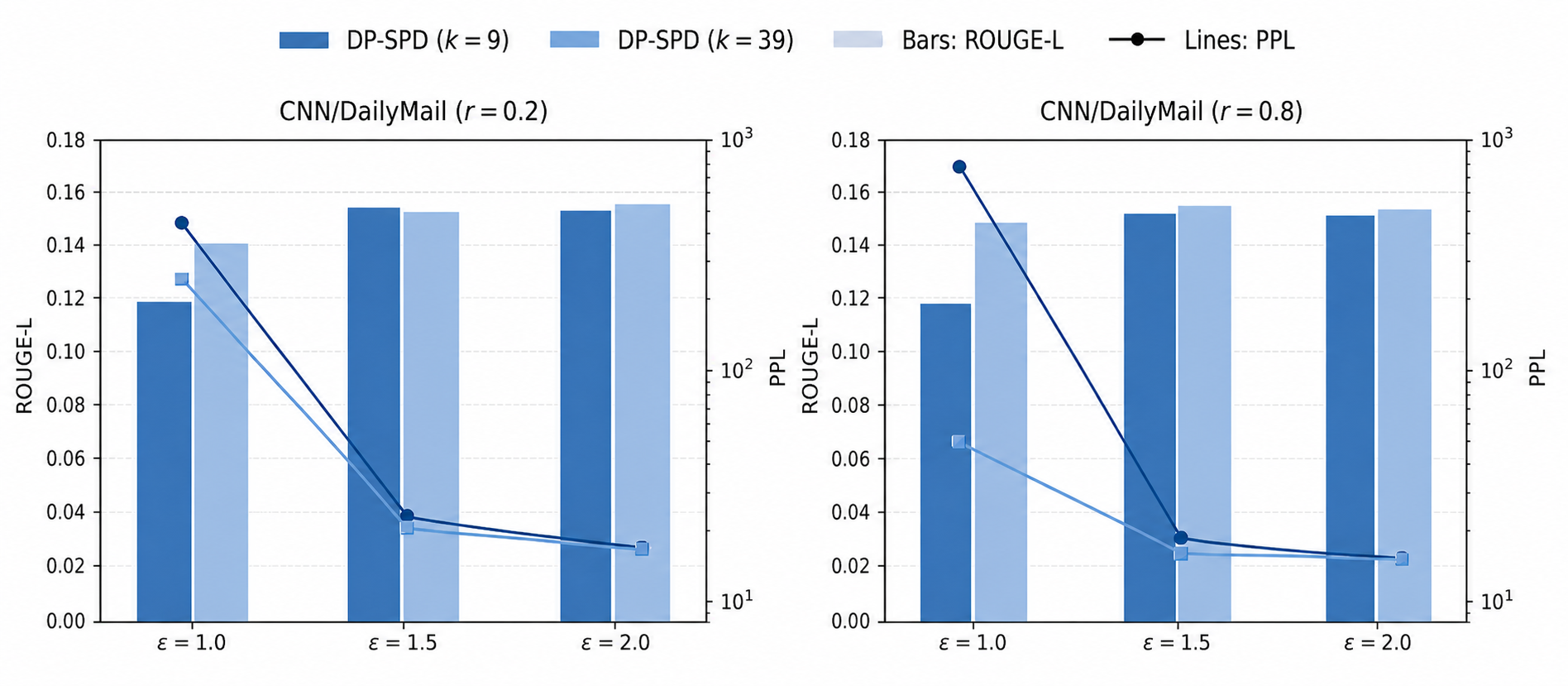}
        \caption{Qwen3-14B on CNN/DailyMail.}
        \label{fig:gen2_appendix}
    \end{subfigure}
    \caption{Generation-utility visualizations moved from the main text to the appendix.}
    \label{fig:generation_appendix}
\end{figure*}

\subsubsection{Qwen3-14B on CNN/DailyMail}

We retain the complete DP-SPD grid over $\epsilon\in\{1.0,1.5,2.0\}$, $r\in\{0.2,0.4,0.6,0.8\}$, and $k\in\{9,19,29,39\}$, followed by the unified related-work comparison.

\begin{table*}[t!]
\centering
\caption{Generation performance of Qwen3-14B on CNN/DailyMail summarization. We report PPL and ROUGE-L under different privacy budgets, dummy ratios, and dummy numbers. Lower PPL and higher ROUGE-L are better.}
\label{tab:qwen3_cnndm_appendix}
\resizebox{0.7\linewidth}{!}{
\begin{tabular}{c|c|cccc|cccc}
\toprule
\multirow{2}{*}{$\epsilon$} & \multirow{2}{*}{$r$}
& \multicolumn{4}{c|}{\textbf{PPL} $\downarrow$}
& \multicolumn{4}{c}{\textbf{ROUGE-L} $\uparrow$} \\
& & $k=9$ & $k=19$ & $k=29$ & $k=39$
& $k=9$ & $k=19$ & $k=29$ & $k=39$ \\
\midrule
1.0 & 0.2 & 253.50 & 161.76 & 141.97 & 137.18 & 0.1197 & 0.1361 & 0.1391 & 0.1411 \\
1.0 & 0.4 & 315.52 & 128.56 & 97.74  & 90.92  & 0.1173 & 0.1387 & 0.1388 & 0.1403 \\
1.0 & 0.6 & 403.63 & 98.93  & 64.93  & 58.43  & 0.1209 & 0.1396 & 0.1433 & 0.1469 \\
1.0 & 0.8 & 553.70 & 71.02  & 40.13  & 35.13  & 0.1184 & 0.1431 & 0.1507 & 0.1504 \\
\midrule
1.5 & 0.2 & 16.07 & 16.04 & 16.05 & 16.05 & 0.1545 & 0.1541 & 0.1526 & 0.1531 \\
1.5 & 0.4 & 15.51 & 15.49 & 15.48 & 15.48 & 0.1533 & 0.1511 & 0.1514 & 0.1513 \\
1.5 & 0.6 & 14.90 & 14.86 & 14.85 & 14.84 & 0.1507 & 0.1497 & 0.1524 & 0.1503 \\
1.5 & 0.8 & 14.18 & 14.12 & 14.11 & 14.11 & 0.1526 & 0.1550 & 0.1548 & 0.1555 \\
\midrule
2.0 & 0.2 & 13.33 & 13.33 & 13.33 & 13.33 & 0.1535 & 0.1551 & 0.1542 & 0.1556 \\
2.0 & 0.4 & 13.28 & 13.28 & 13.28 & 13.28 & 0.1533 & 0.1545 & 0.1535 & 0.1549 \\
2.0 & 0.6 & 13.21 & 13.21 & 13.21 & 13.21 & 0.1555 & 0.1552 & 0.1546 & 0.1543 \\
2.0 & 0.8 & 13.13 & 13.13 & 13.13 & 13.12 & 0.1526 & 0.1538 & 0.1540 & 0.1544 \\
\bottomrule
\end{tabular}
}
\end{table*}

\begin{table*}[t]
\centering
\small
\caption{CNN/DailyMail generation comparison with Qwen3-14B. Each entry is PPL / ROUGE-L. DP-SPD uses $r=0.8,k=39$.}
\label{tab:cnndm_adapter_compare}
\setlength{\tabcolsep}{5pt}
\begin{tabular}{lccc}
\toprule
\textbf{Method} & $\boldsymbol{\epsilon=1.0}$ & $\boldsymbol{\epsilon=1.5}$ & $\boldsymbol{\epsilon=2.0}$ \\
\midrule
\textbf{DP-SPD} & \textbf{35.13 / .1504} & \textbf{14.11 / .1555} & \textbf{13.12 / .1544} \\
CAPE-style & 118680.03 / .0030 & 207624.46 / .0028 & 200676.06 / .0045 \\
CUSTEXT-style & 2897.53 / .0288 & 2606.63 / .0352 & 2297.36 / .0400 \\
DP-Forward-style & 285789.84 / .1264 & 240441.63 / .1314 & 228022.08 / .1318 \\
SANTEXT-style & 18150.13 / .0524 & 18081.09 / .0500 & 18139.80 / .0553 \\
TextObfuscator-style & 25.68 / .1351 & 16.88 / .1499 & 14.65 / .1533 \\
\bottomrule
\end{tabular}
\end{table*}

The InferDPT-style extractor conditional PPLs are 36.29, 36.37, and 36.28 for $\epsilon=1.0,1.5,2.0$. At $\epsilon=1.0$, TextObfuscator-style obtains lower PPL than DP-SPD (25.68 vs. 35.13), while DP-SPD obtains higher ROUGE-L (.1504 vs. .1351); at $\epsilon\in\{1.5,2.0\}$, DP-SPD obtains both lower comparable PPL and slightly higher ROUGE-L.

\subsubsection{Real-world WildChat Utility}

We evaluate Qwen3-32B on 1,000 independent WildChat queries: 500 answerable QA requests and 500 open-ended requests, with target $\epsilon=3$. QA utility is evaluated by semantic correctness against a reference answer, while open-ended responses are scored from 1 to 5 for overall response quality.

The reported overall ROUGE-L is approximately 0.241 for both plaintext and DP-SPD. The QA judge accepts semantic paraphrases and equivalent concise answers; for open-ended prompts, it assigns a 1--5 quality score based primarily on the original user request.

\subsubsection{Style-adapter Implementation Details}

Tables~\ref{tab:adapter_details} shows the details.

\begin{table*}[t]
\centering
\small
\caption{Style-adapted baselines used in the unified LLaMA/Qwen evaluation harness. These are controlled adaptations of the core privacy mechanisms rather than exact reproductions of every original training pipeline.}
\label{tab:adapter_details}
\setlength{\tabcolsep}{4pt}
\begin{tabular}{p{2.4cm}p{5.0cm}p{5.3cm}}
\toprule
\textbf{Adapter} & \textbf{Preserved core mechanism} & \textbf{Adaptation / omitted component} \\
\midrule
CAPE-style & Min--max normalization followed by Laplace perturbation & The original private-attribute adversary and task-head training are not reproduced in the unified harness. \\
CUSTEXT-style & Top-$k$ exponential token mechanism & Original word-vector reconstruction is replaced by the LLaMA/Qwen BPE vocabulary and the common evaluator. \\
DP-Forward-style & Hidden-state perturbation & Full noisy fine-tuning from the original pipeline is not reproduced for the unified classification/generation evaluator. \\
InferDPT-style & RANTEXT-style perturbation and two-stage data flow & IMDb probing does not reproduce the complete remote-LLM/local-extractor pipeline; generation PPL uses an extractor-side conditional definition and is therefore not directly ranked against server-side PPL. \\
SANTEXT-style & Exponential token sanitization & The mechanism is adapted to the tokenizer/BPE vocabulary of the current target model and the unified evaluator. \\
TextObfuscator-style & Client-boundary hidden-state perturbation & Full joint training, prototype losses, and other task-specific training components are not reproduced. \\
\bottomrule
\end{tabular}
\end{table*}

\subsection{Privacy Evaluation Details}
\label{sec:privacy_appendix}

\subsubsection{Query Identification Attack (QIA)}

We use the term \emph{Query Identification Attack (QIA)} for the attack previously referred to as membership inference in earlier drafts. The adversary observes the shuffled set of $k+1$ candidates and attempts to identify the real query. Tables~\ref{tab:qia_imdb_appendix} and~\ref{tab:qia_wikitext_appendix} provide the full dummy-strategy breakdown over $r\in\{20\%,50\%,80\%\}$ and $k\in\{9,19,29\}$, averaged over $\epsilon\in\{1.5,2.0,2.5,3.0\}$.

\begin{table*}[t!]
\centering
\caption{Detailed QIA results on IMDb. Results are averaged over $\epsilon\in\{1.5,2.0,2.5,3.0\}$. Lower ASR and lower PPL gap are better.}
\label{tab:qia_imdb_appendix}
\resizebox{0.8\linewidth}{!}{
\begin{tabular}{l|c|ccc|ccc}
\toprule
\multirow{2}{*}{\textbf{Dummy Strategy}} & \multirow{2}{*}{$r$}
& \multicolumn{3}{c|}{\textbf{QIA ASR (\%)} $\downarrow$}
& \multicolumn{3}{c}{\textbf{PPL Gap} $\downarrow$} \\
& & $k=9$ & $k=19$ & $k=29$ & $k=9$ & $k=19$ & $k=29$ \\
\midrule
DP-SPD & 20\% & 26.50 & 22.50 & 21.00 & 7.41 & 7.38 & 7.42 \\
DP-SPD & 50\% & 13.50 & 10.25 & 8.00  & 8.39 & 8.16 & 8.11 \\
DP-SPD & 80\% & 7.75  & 6.00  & 6.25  & 6.89 & 6.08 & 6.53 \\
\midrule
Random-Sampled & 20\% & 54.75 & 48.00 & 39.50 & 19.40 & 19.59 & 19.96 \\
Random-Sampled & 50\% & 42.00 & 31.75 & 31.25 & 22.14 & 21.64 & 21.03 \\
Random-Sampled & 80\% & 32.25 & 25.75 & 21.00 & 18.09 & 19.92 & 18.25 \\
\midrule
Random-Token & 20\% & 95.75 & 95.75 & 93.50 & 90.64 & 89.96 & 90.44 \\
Random-Token & 50\% & 99.25 & 99.75 & 99.00 & 629.69 & 621.41 & 627.56 \\
Random-Token & 80\% & 99.75 & 99.00 & 99.75 & 3114.79 & 3010.73 & 3002.47 \\
\bottomrule
\end{tabular}
}
\end{table*}

\begin{table*}[t!]
\centering
\caption{Detailed QIA results on WikiText-2. Results are averaged over $\epsilon\in\{1.5,2.0,2.5,3.0\}$. Lower ASR and lower PPL gap are better.}
\label{tab:qia_wikitext_appendix}
\resizebox{0.8\linewidth}{!}{
\begin{tabular}{l|c|ccc|ccc}
\toprule
\multirow{2}{*}{\textbf{Dummy Strategy}} & \multirow{2}{*}{$r$}
& \multicolumn{3}{c|}{\textbf{QIA ASR (\%)} $\downarrow$}
& \multicolumn{3}{c}{\textbf{PPL Gap} $\downarrow$} \\
& & $k=9$ & $k=19$ & $k=29$ & $k=9$ & $k=19$ & $k=29$ \\
\midrule
DP-SPD & 20\% & 24.00 & 17.75 & 13.50 & 16.26 & 15.49 & 14.89 \\
DP-SPD & 50\% & 18.00 & 13.75 & 12.25 & 6.86  & 7.24  & 6.08 \\
DP-SPD & 80\% & 4.00  & 1.50  & 2.25  & 72.95 & 69.94 & 69.72 \\
\midrule
Random-Sampled & 20\% & 54.75 & 42.50 & 31.75 & 99.46  & 98.86  & 96.29 \\
Random-Sampled & 50\% & 36.25 & 29.25 & 25.75 & 278.44 & 259.27 & 243.80 \\
Random-Sampled & 80\% & 17.50 & 13.00 & 8.75  & 519.25 & 542.04 & 513.41 \\
\midrule
Random-Token & 20\% & 78.00 & 74.00 & 67.50 & 316.59  & 307.21  & 306.04 \\
Random-Token & 50\% & 82.50 & 78.75 & 74.50 & 1490.54 & 1518.27 & 1512.71 \\
Random-Token & 80\% & 75.50 & 72.50 & 70.50 & 3480.72 & 3528.30 & 3586.50 \\
\bottomrule
\end{tabular}
}
\end{table*}

\subsubsection{Embedding Inversion Attack}

For EIA, the adversary first maps each privatized embedding to its nearest vocabulary token and then performs group-level candidate identification. Results are averaged over $\epsilon\in\{1.5,3.0\}$ and $r\in\{20\%,50\%,80\%\}$. Increasing the anonymity-set size reduces the group-level ASR for all dummy strategies.

\subsubsection{LLM-assisted Semantic Identification and Reconstruction}

The LLM-assisted attack asks GPT-4o-mini both to select the most plausible real candidate and to reconstruct the private query. Table~\ref{tab:gpt_attack} reports the complete benchmark attack metrics averaged over the tested privacy budgets and dummy ratios.

\subsubsection{WildChat Attack Evaluation}

We further evaluate 400 shuffled WildChat candidate groups with likelihood-based QIA and GPT-4o-mini semantic attacks. The candidate-selection random baseline is $1/(k+1)$, i.e., 10\% for $k=9$ and 5\% for $k=19$. This baseline applies to candidate selection and should not be interpreted as the baseline for exact-query reconstruction.

\begin{table}[t]
\centering
\small
\caption{QIA and semantic-reconstruction results on WildChat. Lower values indicate stronger privacy.}
\label{tab:wildchat_attack_appendix}
\setlength{\tabcolsep}{5pt}
\begin{tabular}{lccc}
\toprule
\textbf{Attack Metric} & $\boldsymbol{k=9}$ & $\boldsymbol{k=19}$ & \textbf{Random selection} \\
\midrule
Min-K QIA ASR & 4.00\% & 1.75\% & 10\% / 5\% \\
PPL QIA ASR & 4.25\% & 2.50\% & 10\% / 5\% \\
GPT selection ASR & 20.0\% & 13.0\% & 10\% / 5\% \\
GPT exact-query recovery & 18.25\% & 11.0\% & -- \\
\bottomrule
\end{tabular}
\end{table}

The likelihood-based attacks fall below the random-selection level in these experiments, whereas GPT-based semantic selection retains a measurable advantage over random guessing.

\subsubsection{Representation Robustness}

To test whether API-only behavior is tied to a particular client embedding space, we keep GPT-4o fixed as the downstream backend on MMLU-Pro and change only the client-side projection model. Although the two embedding spaces yield very different token-retention rates, downstream task accuracy differs by only 0.8 percentage points.

\begin{table}[t]
\centering
\small
\caption{Representation-space robustness for API-only DP-SPD on MMLU-Pro. Values are reported as mean $\pm$ standard deviation.}
\label{tab:representation_bias_appendix}
\setlength{\tabcolsep}{5pt}
\begin{tabular}{lccc}
\toprule
\textbf{Projection Model} & \textbf{Dim.} & \textbf{Token Retention} & \textbf{GPT-4o Acc.} \\
\midrule
Qwen2.5-0.5B & 896 & $98.50\pm0.11$\% & $54.8\pm3.03$\% \\
Gemma-2-2B & 2,304 & $77.85\pm0.15$\% & $54.0\pm2.55$\% \\
\midrule
Qwen $-$ Gemma & -- & +20.65 pp & +0.8 pp \\
\bottomrule
\end{tabular}
\end{table}

\subsection{Efficiency Evaluation Details}
\label{sec:efficiency_appendix}

\subsubsection{Continuation-model Scale}

We vary only the client-side continuation generator among Qwen2.5-0.5B, 3B, 7B, and 32B. All other factors are fixed: WikiText-2 test with 300 samples, target $\epsilon=3$, $k=9$, $r=0.2$, $\epsilon_{\mathrm{non\mbox{-}sensitive}}=3.0$, $\epsilon_{\mathrm{private}}=7.82$, Laplace noise with clipping norm 1, Qwen2.5-0.5B-Instruct as the transport/projection model, and Qwen2.5-0.5B-Instruct as the common quality/QIA scorer. Every group contains one real candidate and nine dummies, so random candidate selection is 10\%.

\par\medskip
\noindent\begin{minipage}{\linewidth}
\centering
\captionsetup{font=footnotesize,skip=5pt}
\captionof{table}{Effect of continuation-model scale. Surface ROUGE-L is measured on the final DP-perturbed and projected dummy suffix. Held-out PPL is scored by each tested proxy itself (not server-side generated-text PPL). QIA uses a fixed 0.5B scorer. Bracketed values indicate the reported intervals.}
\label{tab:proxy_scale_appendix}
\vspace{3pt}
{\footnotesize
\setlength{\tabcolsep}{5pt}
\renewcommand{\arraystretch}{1.10}
\textit{(a) Dummy quality and generation time}\par\vspace{3pt}
\begin{tabular}{@{}lrrrr@{}}
\toprule
\textbf{Proxy} & \textbf{Params} & \textbf{Surface R-L $\uparrow$} & \textbf{Held-out PPL $\downarrow$} & \textbf{s/group $\downarrow$} \\
\midrule
0.5B & 0.494B & .1138 [.1067,.1211] & 14.95 [13.83,16.15] & \textbf{0.813} \\
3B   & 3.086B & .1424 [.1332,.1522] & 9.86 [9.16,10.63]   & 1.198 \\
7B   & 7.616B & .1555 [.1450,.1663] & 8.61 [7.97,9.29]    & 1.092 \\
32B  & 32.764B & \textbf{.1691 [.1547,.1841]} & \textbf{6.33 [5.83,6.89]} & 3.756 \\
\bottomrule
\end{tabular}

\vspace{9pt}
\textit{(b) Throughput and query-identification attacks}\par\vspace{3pt}
\begin{tabular}{@{}lrrrr@{}}
\toprule
\textbf{Proxy} & \textbf{Total (s)} & \textbf{tok/s $\uparrow$} & \textbf{PPL-QIA $\downarrow$} & \textbf{Min-K QIA $\downarrow$} \\
\midrule
0.5B & 243.93 & \textbf{321.5} & 8.33\% [5.71,12.01] & \textbf{1.33\% [0.52,3.38]} \\
3B   & 359.35 & 218.2 & \textbf{6.00\% [3.83,9.28]} & 5.00\% [3.05,8.08] \\
7B   & 327.75 & 239.3 & \textbf{6.00\% [3.83,9.28]} & 2.67\% [1.36,5.17] \\
32B  & 1126.69 & 69.6 & 7.33\% [4.89,10.85] & 2.67\% [1.36,5.17] \\
\bottomrule
\end{tabular}
}
\end{minipage}
\par\medskip

Surface ROUGE-L increases monotonically with proxy size. The fixed 0.5B evaluator PPL on the final dummy candidates is 68.26, 67.58, 71.69, and 73.77 for the 0.5B/3B/7B/32B proxies, respectively, and therefore does not improve monotonically. Generation time measures CUDA-synchronized \texttt{proxy\_model.generate} only and excludes CPU DP noise and nearest-token projection. The attack point estimates remain around or below the 10\% random baseline without a monotonic increase as the proxy becomes larger.

\Needspace{14\baselineskip}
\subsubsection{KV-cache Inference-Time Breakdown}

Let $p=1-r$ denote the shared-prefix ratio. The no-sharing baseline uses $p=0$ and processes each candidate independently. Table~\ref{tab:mha_breakdown_complete} provides the complete inference-time breakdown for sequence length 1024. The strongest measured operating point, $p=0.8$ ($r=0.2$) and $k=24$, reduces total inference time from 3868.02 ms to 1002.86 ms, corresponding to a $3.86\times$ speedup.

\par\medskip
\noindent\begin{minipage}{\linewidth}
\centering
\captionsetup{font=footnotesize,skip=5pt}
\captionof{table}{Inference-time breakdown for 1024-token inputs. MHA, linear-layer, and total CUDA times are in milliseconds. Speedups use the no-sharing baseline ($p=0$) at the same dummy count $k$.}
\label{tab:mha_breakdown_complete}
\vspace{3pt}
{\footnotesize
\setlength{\tabcolsep}{5pt}
\renewcommand{\arraystretch}{1.08}
\begin{tabular}{@{}c c r r r c c@{}}
\toprule
\textbf{$p$} & \textbf{$k$} & \textbf{MHA (ms)} & \textbf{Linear (ms)} & \textbf{Total (ms)} & \textbf{MHA gain} & \textbf{Total gain} \\
\midrule
\multirow{5}{*}{0.0 (Naive)}
& 0  & 67.34  & 84.35  & 175.07 & 1.00$\times$ & 1.00$\times$ \\
& 9  & 607.90 & 723.57 & 1527.27 & 1.00$\times$ & 1.00$\times$ \\
& 14 & 899.69 & 1089.36 & 2281.81 & 1.00$\times$ & 1.00$\times$ \\
& 19 & 1216.77 & 1460.89 & 3066.87 & 1.00$\times$ & 1.00$\times$ \\
& 24 & 1526.78 & 1852.62 & 3868.02 & 1.00$\times$ & 1.00$\times$ \\
\midrule
\multirow{5}{*}{0.2}
& 0  & 68.54  & 86.04  & 177.48 & 0.98$\times$ & 0.99$\times$ \\
& 9  & 533.55 & 639.53 & 1346.36 & 1.14$\times$ & 1.13$\times$ \\
& 14 & 788.00 & 929.74 & 1968.51 & 1.14$\times$ & 1.16$\times$ \\
& 19 & 1038.61 & 1229.86 & 2596.67 & 1.17$\times$ & 1.18$\times$ \\
& 24 & 1285.44 & 1520.59 & 3212.09 & 1.19$\times$ & 1.20$\times$ \\
\midrule
\multirow{5}{*}{0.4}
& 0  & 68.66  & 86.15  & 177.68 & 0.98$\times$ & 0.99$\times$ \\
& 9  & 426.36 & 516.64 & 1078.04 & 1.43$\times$ & 1.42$\times$ \\
& 14 & 609.08 & 725.42 & 1527.90 & 1.48$\times$ & 1.49$\times$ \\
& 19 & 814.14 & 949.42 & 2018.03 & 1.49$\times$ & 1.52$\times$ \\
& 24 & 1018.54 & 1171.71 & 2500.22 & 1.50$\times$ & 1.55$\times$ \\
\midrule
\multirow{5}{*}{0.6}
& 0  & 69.10 & 86.83 & 178.84 & 0.97$\times$ & 0.98$\times$ \\
& 9  & 319.80 & 368.66 & 787.81 & 1.90$\times$ & 1.94$\times$ \\
& 14 & 450.98 & 522.17 & 1109.93 & 1.99$\times$ & 2.06$\times$ \\
& 19 & 585.56 & 664.19 & 1425.83 & 2.08$\times$ & 2.15$\times$ \\
& 24 & 723.86 & 805.43 & 1744.20 & 2.11$\times$ & 2.22$\times$ \\
\midrule
\multirow{5}{*}{0.8}
& 0  & 68.68 & 86.19 & 177.69 & 0.98$\times$ & 0.99$\times$ \\
& 9  & 226.49 & 243.16 & 532.95 & 2.68$\times$ & 2.87$\times$ \\
& 14 & 290.66 & 317.05 & 690.63 & 3.10$\times$ & 3.30$\times$ \\
& 19 & 362.52 & 390.01 & 855.83 & 3.36$\times$ & 3.58$\times$ \\
& 24 & 426.51 & 453.97 & 1002.86 & 3.58$\times$ & 3.86$\times$ \\
\bottomrule
\end{tabular}
}
\end{minipage}
\par\medskip

\Needspace{12\baselineskip}
\subsubsection{Privacy--Utility--Efficiency Trade-off}

The two system parameters $k$ and $r$ jointly determine the effective perturbation level and server-side cost. Increasing $k$ strengthens shuffle amplification, while increasing the dummy ratio $r$ protects a larger portion of the input but shortens the reusable prefix. The complete trade-off is visualized in Figure~\ref{fig:3d}. At comparable noise levels, a small dummy ratio (e.g., $r=0.2$, hence $p=0.8$) can achieve more than a 3.2× inference speedup while maintaining similar privacy protection.

\par\medskip
\noindent\begin{minipage}{\linewidth}
\centering
\includegraphics[width=0.96\linewidth]{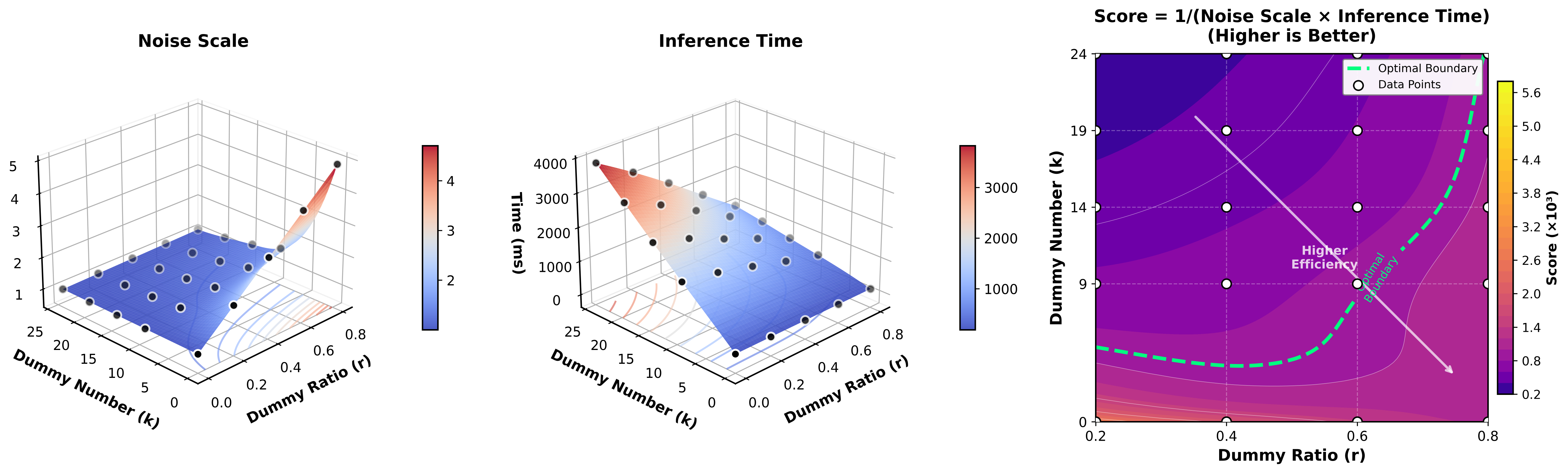}
\captionsetup{font=footnotesize,skip=4pt}
\captionof{figure}{Trade-off between noise scale and inference time across dummy ratios $r$ and dummy numbers $k$.}
\label{fig:3d}
\end{minipage}
\par\medskip

The KV-cache efficiency evaluation also shows the same trend at $r=0.6$: for $k=39$, inference time decreases from 2919.36 ms without prefix reuse to 1906.68 ms with shared-prefix reuse, a 34.7\% reduction, while the associated classification-accuracy change is below 0.25 percentage points in the matched comparison. These results illustrate that KV-cache reuse can offset a substantial fraction of the compute overhead introduced by larger anonymity sets.

\section{Prompt for GPT Inference Attack}
\label{sec:gpt_attack}

\begin{tcolorbox}[
    colback=gray!4,
    colframe=gray!55,
    boxrule=0.6pt,
    arc=2mm,
    left=1mm,
    right=1mm,
    top=1mm,
    bottom=1mm,
    title={Prompt used for GPT inference attack},
    fonttitle=\bfseries,
    breakable
]
\begin{Verbatim}[fontsize=\small, breaklines=true, breakanywhere=true]
You are performing a GPT inference attack adapted from InferDPT.

Context:
- A privacy-preserving inference system produced one group of candidate queries.
- Exactly one candidate is the real user query.
- The other candidates are dummy continuations generated from the same shared prefix.
- The text you see below was reconstructed from noised embeddings, so it may contain token errors.

Your tasks:
1. Decide which candidate suffix is most likely to come from the real query.
2. Recover the most likely original full query before noise.
3. Briefly explain which semantic or stylistic cues made that candidate more plausible than the others.

Return valid JSON only in the following schema:
{
  "predicted_real_index": 0,
  "recovered_query": "full recovered query here",
  "confidence": 0.0,
  "brief_reasoning": "short explanation"
}

Shared prefix:
This was a quite brutalFileSystem. There were huge implausibilities, and a silly script, bad acting, etc.<br /><br />The only reason to watch this movie is that from-time to

Candidate suffixes:
Candidate 0 suffix:
 time a person mentions "Dyslexia Syndrome

Candidate 1 suffix:
 time you hear the words "The most expensive movie

Candidate 2 suffix:
 time it reminds you how wonderful a film Director can

Candidate 3 suffix:
 time someone will askwhy I have no internet connection

Candidate 4 suffix:
 time it shows on TV or through internet streaming services

Candidate 5 suffix:
 time an audience might have noticed that it was really

Candidate 6 suffix:
 time they had a little scene which objectives how good

Candidate 7 suffix:
time the narrator of the scene makes the film look

Candidate 8 suffix:
 time some quite impressive sets of breasts were exposed.

Candidate 9 suffix:
 time it makes me laugh because of how ridiculous one
\end{Verbatim}
\end{tcolorbox}

\end{document}